\documentclass{article}

\usepackage[nonatbib, final]{neurips_2020}

\usepackage[utf8]{inputenc} % allow utf-8E input
\usepackage[T1]{fontenc}    % use 8-bit T1 fonts
\usepackage{hyperref}       % hyperlinks
\usepackage{url}            % simple URL typesetting
\usepackage{booktabs}       % professional-quality tables
\usepackage{amsfonts}       % blackboard math symbols
\usepackage{nicefrac}       % compact symbols for 1/2, etc.
\usepackage{microtype}      % microtypography
\usepackage{xcolor}
\usepackage{graphicx}
\usepackage[square,numbers]{natbib}

\usepackage{algorithm}
\usepackage[noend]{algpseudocode}

\usepackage{amsmath}
\usepackage{svg}
\usepackage{import}
\usepackage{comment}
\usepackage{graphicx,wrapfig,lipsum}
\usepackage[binary-units=true]{siunitx}
\usepackage{textcomp}

\newcommand{\norm}[1]{\left\lVert#1\right\rVert}
\DeclareMathOperator*{\argmin}{arg\,min}

\DeclareMathOperator{\SO}{SO}

\usepackage{scalerel,stackengine}
\stackMath
\newcommand\reallywidehat[1]{%
\savestack{\tmpbox}{\stretchto{%
  \scaleto{%
    \scalerel*[\widthof{\ensuremath{#1}}]{\kern-.6pt\bigwedge\kern-.6pt}%
    {\rule[-\textheight/2]{1ex}{\textheight}}%WIDTH-LIMITED BIG WEDGE
  }{\textheight}% 
}{0.5ex}}%
\stackon[1pt]{#1}{\tmpbox}%
}

\newcommand*\samethanks[1][\value{footnote}]{\footnotemark[#1]}

\title{Neural Non-Rigid Tracking}

\author{%
  Alja{\v{z}} Bo{\v{z}}i{\v{c}}$^{1}$\thanks{Denotes equal contribution.}\\
  \texttt{aljaz.bozic@tum.de}
  \And
  Pablo Palafox$^{1}$\samethanks\\
  \texttt{pablo.palafox@tum.de}
  \AND
  Michael Zollh{\"o}fer$^{2}$
  \And
  Angela Dai$^{1}$
  \And
  Justus Thies$^{1}$
  \And
  Matthias Nie{\ss}ner$^{1}$
  \AND
  {\normalfont $^{1}$Technical University of Munich}
  \And
  {\normalfont $^{2}$Stanford University}
}

\begin{document}

\maketitle

\begin{abstract}

We introduce a novel, end-to-end learnable, differentiable non-rigid tracker that enables state-of-the-art non-rigid reconstruction by a learned robust optimization.
Given two input RGB-D frames of a non-rigidly moving object, we employ a convolutional neural network to predict dense correspondences and their confidences.
These correspondences are used as constraints in an as-rigid-as-possible (ARAP) optimization problem.
By enabling gradient back-propagation through the weighted non-linear least squares solver, we are able to learn correspondences and  confidences in an end-to-end manner such that they are optimal for the task of non-rigid tracking.
Under this formulation, correspondence confidences can be learned via self-supervision, informing a learned robust optimization, where outliers and wrong correspondences are automatically down-weighted to enable effective tracking.
Compared to state-of-the-art approaches, our algorithm shows improved reconstruction performance, while simultaneously achieving $85\times$ faster correspondence prediction than comparable deep-learning based methods.
We make our code available at \url{https://github.com/DeformableFriends/NeuralTracking}.
\end{abstract}
\section{Introduction}

%%%%%%%%%%%%%%%%%%%%%%%%%%%%%%%%%%%%%%%%%%%%%%%%%%%%%%%%%%%%%%%%%%%%%%%%%%%%%%%%%%%
% Importance of non-rigid reconstruction using single RGB-D camera
%%%%%%%%%%%%%%%%%%%%%%%%%%%%%%%%%%%%%%%%%%%%%%%%%%%%%%%%%%%%%%%%%%%%%%%%%%%%%%%%%%%
The capture and reconstruction of real-world environments is a core problem in computer vision, enabling numerous VR/AR applications.
While there has been significant progress in reconstructing static scenes, tracking and reconstruction of dynamic objects remains a challenge.
Non-rigid reconstruction focuses on dynamic objects, without assuming any explicit shape priors, such as human or face parametric models.
Commodity \mbox{RGB-D} sensors, such as Microsoft's Kinect or Intel's Realsense, provide a cost-effective way to acquire both color and depth video of dynamic motion.
Using a large number of \mbox{RGB-D} sensors can lead to an accurate non-rigid reconstruction, as shown by~\citet{dou2016fusion4d}.
Our work focuses on non-rigid reconstruction from a single \mbox{RGB-D} camera, thus eliminating the need for specialized multi-camera setups.
%

%%%%%%%%%%%%%%%%%%%%%%%%%%%%%%%%%%%%%%%%%%%%%%%%%%%%%%%%%%%%%%%%%%%%%%%%%%%%%%%%%%%
%%%%%%%%%%%%%%%%%%%%%%%%%%%%%%%%%%%%%%%%%%%%%%%%%%%%%%%%%%%%%%%%%%%%%%%%%%%%%%%%%%%
The seminal DynamicFusion by \citet{newcombe2015dynamicfusion} introduced a non-rigid tracking and mapping pipeline that uses depth input for real-time non-rigid reconstruction from a single \mbox{RGB-D} camera.
Various approaches have expanded upon this framework by incorporating sparse color correspondences~\cite{innmann2016volumedeform} or dense photometric optimization~\cite{guo2017real}.
DeepDeform~\cite{bozic2020deepdeform} presented a learned correspondence prediction, enabling significantly more robust tracking of fast motion and re-localization.
Unfortunately, the computational cost of the correspondence prediction network ($\sim 2$ seconds per frame for a relatively small number of non-rigid correspondences) inhibits real-time performance.

Simultaneously, work on learned optical flow has shown dense correspondence prediction at real-time rates~\cite{sun2018pwc}.
However, directly replacing the non-rigid correspondence predictions from \citet{bozic2020deepdeform} with these optical flow predictions does not produce accurate enough correspondences for comparable non-rigid reconstruction performance. 
In our work, we propose a neural non-rigid tracker, i.e., an end-to-end differentiable non-rigid tracking pipeline which combines the advantages of classical deformation-graph-based reconstruction pipelines~\cite{newcombe2015dynamicfusion, innmann2016volumedeform} with novel learned components.
Our end-to-end approach enables learning outlier rejection in a self-supervised manner, which guides a robust optimization to mitigate the effect of inaccurate correspondences or major occlusions present in single RGB-D camera scenarios.

Specifically, we cast the non-rigid tracking problem as an as-rigid-as-possible (ARAP) optimization problem, defined on correspondences between points in a source and a target frame.
A differentiable Gauss-Newton solver allows us to obtain gradients that enable training a neural network to predict an importance weight for every correspondence in a completely self-supervised manner, similar to robust optimization.
The end-to-end training significantly improves non-rigid tracking performance. Using our neural tracker in a non-rigid reconstruction framework results in $85\times$ faster correspondence prediction and improved reconstruction performance compared to the state of the art.

%%%%%%%%%%%%%%%%%%%%%%%%%%%%%%%%%%%%%%%%%%%%%%%%%%%%%%%%%%%%%%%%%%%%%%%%%%%%%%%%%%%
% Contribution summary
%%%%%%%%%%%%%%%%%%%%%%%%%%%%%%%%%%%%%%%%%%%%%%%%%%%%%%%%%%%%%%%%%%%%%%%%%%%%%%%%%%%
In summary, we propose a novel neural non-rigid tracking approach with two key contributions:

\begin{itemize}
    \item an end-to-end differentiable Gauss-Newton solver, which provides gradients to better inform a correspondence prediction network used for non-rigid tracking of two frames;
    \item a self-supervised approach for learned correspondence weighting, which is informed by our differentiable solver and enables efficient, robust outlier rejection, thus, improving non-rigid reconstruction performance compared to the state of the art.
\end{itemize}

\begin{figure*}
\centering{
    \def\svgwidth{1\linewidth}
    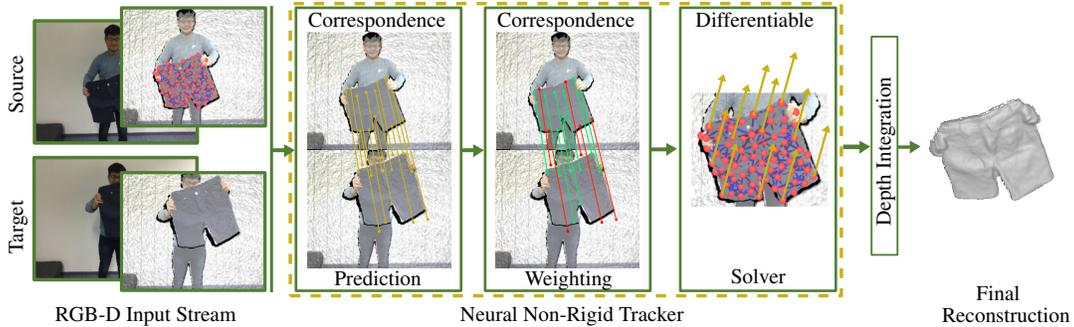
    \caption{Neural Non-Rigid Tracking: based on RGB-D input data of a source and a target frame, our learned non-rigid tracker estimates the non-rigid deformations to align the source to the target frame. 
    We propose an end-to-end approach, enabling correspondences and their importance weights to be informed by the non-rigid solver. Similar to robust optimization, this provides robust tracking, and the resulting deformation field can then be used to integrate the depth observations in a canonical volumetric 3D grid that implicitly represents the surface of the object (final reconstruction).}
    \label{fig:teaser}
}
\end{figure*}
\section{Related Work}

\paragraph{Non-rigid Reconstruction.}
Reconstruction of deformable objects using a single RGB-D camera is an important research area in computer vision.
State-of-the-art methods rely on deformation graphs \cite{Sumner2007ed, zollhoefer2014deformable} that enable robust and temporally consistent 3D motion estimation.
While earlier approaches required an object template, such graph-based tracking has been extended to simultaneous tracking and reconstruction approaches~\cite{dou20153d, newcombe2015dynamicfusion}.
These works used depth fitting optimization objectives in the form of iterative closest points, or continuous depth fitting in \cite{slavcheva2017killingfusion, slavcheva2018sobolevfusion}.
Rather than relying solely on depth information, recent works have incorporated SIFT features 
\cite{innmann2016volumedeform}, dense photometric fitting \cite{guo2017real}, and sparse learned correspondence \cite{bozic2020deepdeform}.

\paragraph{Correspondence Prediction for Non-rigid Tracking.}
In non-rigid tracking, correspondences must be established between the two frames we want to align.
While methods such as DynamicFusion~\cite{newcombe2015dynamicfusion} rely on projective correspondences, recent methods leverage learned correspondences~\cite{bozic2020deepdeform}.
DeepDeform~\cite{bozic2020deepdeform} relies on sparse predicted correspondences, trained on an annotated dataset of deforming objects.
Since prediction is done independently for each correspondence, this results in a high compute cost, compared to dense predictions of state-of-the-art optical flow networks.
Optical flow~\cite{dosovitskiy2015flownet, ilg2017flownet2, sun2018pwc, liu2019selflow} and scene flow~\cite{ma2019deep, behl2019pointflownet, liu2019flownet3d, wang2020flownet3d++} methods achieve promising results in predicting dense correspondences between two frames, with some approaches not even requiring direct supervision \cite{wang2019learning, li2019joint, lai2020mast}.
In our proposed neural non-rigid tracking approach, we build upon PWC-Net~\cite{sun2018pwc} for dense correspondence prediction to inform our non-rigid deformation energy formulation.
Since our approach allows for end-to-end training, our 2D correspondence prediction finds correspondences better suited for non-rigid tracking.

\paragraph{Differentiable Optimization.}
Differentiable optimizers have been explored for various tasks, including image alignment~\cite{chang2017clkn}, rigid pose estimation~\cite{han2018regnet, lv2019taking}, multi-frame direct bundle-adjustment~\cite{tang2018ba}, and rigid scan-to-CAD alignment~\cite{avetisyan2019end}.
In addition to achieving higher accuracy, an end-to-end differentiable optimization approach also offers the possibility to optimize run-time, as demonstrated by learning efficient pre-conditioning methods in \cite{gotz2018machine, sappl2019deepprecond, li2020learning}.
Unlike Li et al. \cite{li2020learning}, which employs an image-based tracker (with descriptors defined on nodes in a pixel-aligned graph), our approach works on general graphs and learns to robustify correspondence prediction for non-rigid tracking by learning self-supervised correspondence confidences.
\section{Non-Rigid Reconstruction Notation} \label{sec:notation}

Non-rigid alignment is a crucial part of non-rigid reconstruction pipelines.
In the single RGB-D camera setup, we are given a pair of source and target RGB-D frames $\{ (\mathcal{I}_s, \mathcal{P}_s), (\mathcal{I}_t, \mathcal{P}_t)  \}$, where $\mathcal{I}_* \in \mathbb{R}^{H \times W \times 3}$ is an RGB image and $\mathcal{P}_* \in \mathbb{R}^{H \times W \times 3}$ a 3D point image.
The goal is to estimate a warp field $\mathcal{Q} : \mathbb{R}^3 \mapsto \mathbb{R}^3$ that transforms $\mathcal{P}_s$ into the target frame.
Note that we define the 3D point image $\mathcal{P}_s$ as the result of back-projecting every pixel $\mathbf{u} \in \Pi_s \subset \mathbb{R}^2$ into the camera coordinate system with given camera intrinsic parameters.
To this end, we define the inverse of the perspective projection to back-project a pixel $\mathbf{u}$ given the pixel's depth $d_\mathbf{u}$ and the intrinsic camera parameters $\mathbf{c}$:
\begin{equation}
\pi^{-1}_{\mathbf{c}} : \mathbb{R}^2 \times \mathbb{R} \rightarrow \mathbb{R}^3, \quad (\mathbf{u}, d_\mathbf{u}) \mapsto \pi^{-1}_{\mathbf{c}}(\mathbf{u}, d_\mathbf{u}) = \mathbf{p}.    
\label{eq:backproject}
\end{equation}
%
%% Deformation graph
To maintain robustness against noise in the depth maps, state-of-the-art approaches define an embedded deformation graph $\mathcal{G} = \{ \mathcal{V}, \mathcal{E} \}$ over the \emph{source} RGB-D frame, where $\mathcal{V}$ is the set of graph nodes defined by their 3D coordinates $\mathbf{v}_i \in \mathbb{R}^3$ and $\mathcal{E}$ the set of edges between nodes, as described in~\cite{Sumner2007ed} and illustrated in Fig.~\ref{fig:teaser}.
Thus, for every node in $\mathcal{G}$, a global translation vector $\mathbf{t}_{\mathbf{v}_i} \in \mathbb{R}^3$ and a rotation matrix $\mathbf{R}_{\mathbf{v}_i} \in \mathbb{R}^{3\times3}$, must be estimated in the alignment process.
We parameterize rotations with a 3-dimensional axis-angle vector $\boldsymbol{\omega} \in \mathbb{R}^3$.
We use the exponential map $\exp : \mathfrak{so}(3)\rightarrow \SO(3), \quad \widehat{\hspace{0pt}\boldsymbol{\omega}\hspace{0pt}} \mapsto \mathrm{e}^{\widehat{\hspace{0pt}\boldsymbol{\omega}\hspace{0pt}}} = \mathbf{R}$ to convert from axis-angle to matrix rotation form, where the~$\, \widehat{\cdot}$-operator creates a $3 \times 3$ skew-symmetric matrix from a 3-dimensional vector. 
The resulting graph motion is denoted by $\mathcal{T} = (\boldsymbol{\omega}_{\mathbf{v}_1}, \mathbf{t}_{\mathbf{v}_1}, \dots, \boldsymbol{\omega}_{\mathbf{v}_N}, \mathbf{t}_{\mathbf{v}_N}) \in \mathbb{R}^{N \times 6}$ for a graph with $N$ nodes.

%%% Warp operation
Dense motion can then be computed by interpolating the nodes' motion $\mathcal{T}$ by means of a warping function $\mathrm{Q}$.
When applied to a 3D point $\mathbf{p} \in \mathbb{R}^{3}$, it produces the point's deformed position
\begin{equation}
    \mathrm{Q} (\mathbf{p}, \mathcal{T}) = \sum_{\mathbf{v}_i \in \mathcal{V}}{ \alpha_{\mathbf{v}_i} ( \mathrm{e}^{\widehat{\hspace{0pt}\boldsymbol{\omega}\hspace{0pt}}_{\mathbf{v}_i}} (\mathbf{p} - \mathbf{v}_i) + \mathbf{v}_i + \mathbf{t}_{\mathbf{v}_i}) }.
    \label{eq:Q}
\end{equation}
The weights $\alpha_{\mathbf{v}_i} \in \mathbb{R}$, also known as \emph{skinning} weights, measure the influence of each node on the current point $\mathbf{p}$ and are computed as in \cite{DoubleFusion}. Please see the supplemental material for further detail.
\section{Neural Non-rigid Tracking}

\begin{figure*}
\centering{
    \def\svgwidth{1\linewidth}
    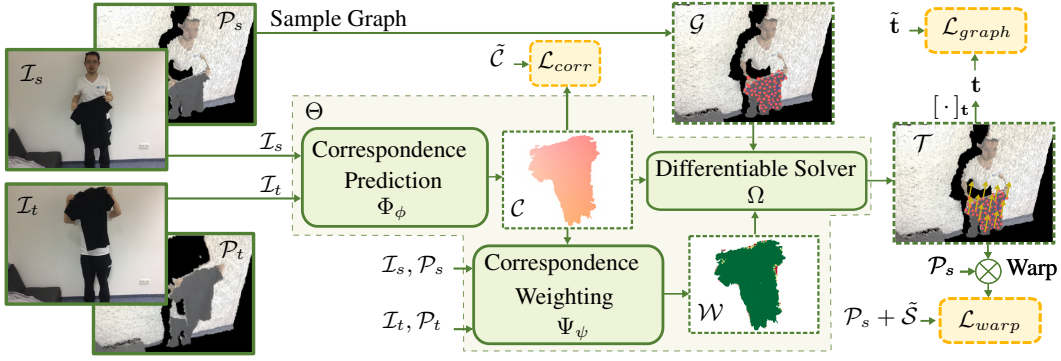
  \caption{Overview of our neural non-rigid tracker.
  Given a pair of source and target images, $\mathcal{I}_s$~and~$\mathcal{I}_t$, a dense correspondence map $\mathcal{C}$ between the frames is estimated via a convolutional neural network~$\Phi$.
  Importance weights $\mathcal{W}$ for these correspondences are computed through a function~$\Psi$.
  Together with a graph $\mathcal{G}$ defined over the source RGB-D frame $\mathcal{P}_s$, both $\mathcal{C}$ and $\mathcal{W}$ are input to a differentiable solver $\Omega$.
  The solver outputs the graph motion $\mathcal{T}$, i.e., the non-rigid alignment between source and target frames.
  Our approach is optimized end-to-end, with losses on the final alignment using $\mathcal{L}_{\mathrm{graph}}$ and $\mathcal{L}_\mathrm{warp}$, and an intermediate loss on the correspondence map $\mathcal{L}_\mathrm{corr}$.
  }
  \label{fig:network_overview}
}
\end{figure*}

Given a pair of source and target RGB-D frames $\left(\mathcal{Z}_s, \mathcal{Z}_t \right)$, where $ \mathcal{Z}_* = (\mathcal{I}_* | \mathcal{P}_*) \in \mathbb{R}^{H \times W \times 6}$ is the concatenation of an RGB and a 3D point image as defined in Section~\ref{sec:notation}, we aim to find a function~$\Theta$ that estimates the motion $\mathcal{T}$ of a deformation graph $\mathcal{G}$ with $N$ nodes (given by their 3D coordinates~$\mathcal{V}$) defined over the source RGB-D frame.
This implicitly defines source-to-target dense 3D motion (see Figure~\ref{fig:network_overview}).
Formally, we have:
\begin{equation}
\Theta: \mathbb{R}^{H \times W \times 6} \times \mathbb{R}^{H \times W \times 6} \times \mathbb{R}^{N \times 3}  \rightarrow \mathbb{R}^{N \times 6}, \quad \left(\mathcal{Z}_s, \mathcal{Z}_t, \mathcal{V}\right) \mapsto \Theta\left(\mathcal{Z}_s, \mathcal{Z}_t, \mathcal{V}\right)=\mathcal{T}.
\end{equation}
To estimate $\mathcal{T}$, we first establish dense 2D correspondences between the source and target frame using a deep neural network $\Phi$. 
These correspondences, denoted as $\mathcal{C}$, are used to construct the data term in our non-rigid alignment optimization.
Since the presence of outlier correspondence predictions has a strong negative impact on the performance of non-rigid tracking, we introduce a weighting function~$\Psi$, inspired by robust optimization, to down-weight inaccurate predictions. 
Function $\Psi$ outputs importance weights $\mathcal{W}$ and is learned in a self-supervised manner.
Finally, both correspondence predictions $\mathcal{C}$ and importance weights $\mathcal{W}$ are input to a differentiable, non-rigid alignment optimization module~$\Omega$.
By optimizing the non-rigid alignment energy (see Section~\ref{sec:differentiable_optimizer}), the differentiable optimizer~$\Omega$ estimates the deformation graph parameters $\mathcal{T}$ that define the motion from source to target frame:
\begin{equation}
\mathcal{T} =  \Theta\left(\mathcal{Z}_s, \mathcal{Z}_t, \mathcal{V} \right) = \Omega\left(\Phi(\cdot), \Psi(\cdot), \mathcal{V} \right) = \Omega\left(\mathcal{C}, \mathcal{W}, \mathcal{V} \right).
\end{equation}
In the following, we define the dense correspondence predictor $\Phi$, the importance weighting $\Psi$ and the optimizer $\Omega$, and describe a fully differentiable approach for optimizing $\Phi$ and $\Psi$ such that we can estimate dense correspondences with importance weights best suited for non-rigid tracking.

%%%%%%%%%%%%%%%%%%%%%%%%%%%%%%%%%%%%%%%%%%%%%%%%%%%%%%%%%%%%%%%%%%%%%%%%%%%%%%%%%%%
% Dense Correspondence Prediction
%%%%%%%%%%%%%%%%%%%%%%%%%%%%%%%%%%%%%%%%%%%%%%%%%%%%%%%%%%%%%%%%%%%%%%%%%%%%%%%%%%%
\subsection{Dense Correspondence Prediction}
The dense correspondence prediction function $\Phi$ takes as input a pair of source and target RGB images $(\mathcal{I}_s, \mathcal{I}_t)$, and for each source pixel location $\mathbf{u} \in \Pi_s \subset \mathbb{R}^2$ it outputs a corresponding pixel location in the target image $\mathcal{I}_t$, which we denote by $\mathbf{c}_\mathbf{u} \in \Pi_t \subset \mathbb{R}^2$.
Formally, $\Phi$ is defined as
\begin{equation}
\Phi: \mathbb{R}^{H \times W \times 3} \times \mathbb{R}^{H \times W \times 3}  \rightarrow \mathbb{R}^{H \times W \times 2}, \quad \left( \mathcal{I}_s, \mathcal{I}_t \right) \mapsto \Phi\left( \mathcal{I}_s, \mathcal{I}_t \right)=\mathcal{C},
\label{eq:Phi}
\end{equation}
where $\mathcal{C}$ is the resulting dense correspondence map.
The function $\Phi$ is represented by a deep neural network that leverages the architecture of a state-of-the-art optical flow estimator~\cite{sun2018pwc}.

%%%%%%%%%%%%%%%%%%%%%%%%%%%%%%%%%%%%%%%%%%%%%%%%%%%%%%%%%%%%%%%%%%%%%%%%%%%%%%%%%%%
% Weights
%%%%%%%%%%%%%%%%%%%%%%%%%%%%%%%%%%%%%%%%%%%%%%%%%%%%%%%%%%%%%%%%%%%%%%%%%%%%%%%%%%%
\subsection{Correspondence Importance Weights}
For each source pixel $\mathbf{u} \in  \Pi_s \subset \mathbb{R}^2$ and its correspondence $\mathbf{c}_\mathbf{u} \in  \Pi_t \subset \mathbb{R}^2$, we additionally predict an importance weight $w_{\mathbf{u}} \in (0, 1)$ by means of the weighting function $\Psi$.
The latter takes as input the source RGB-D image $\mathcal{Z}_s$, the corresponding sampled target frame values $\mathcal{Z}'_t$, and intermediate features from the correspondence network $\Phi$, and outputs weights for the correspondences between source and target.
Note that $\mathcal{Z}'_t$ is the result of bilinearly sampling~\cite{jaderberg2015spatial} the target image $\mathcal{Z}_t$ at the predicted correspondence locations $\mathcal{C}$.
The last layer of features $\mathcal{H}$ of the correspondence network $\Phi$, with dimension $D = 565$, are used to inform $\Psi$.
The weighting function is thus defined as
\begin{equation}
\Psi: \mathbb{R}^{H \times W \times 6} \times \mathbb{R}^{H \times W \times 6} \times \mathbb{R}^{H \times W \times D}  \rightarrow \mathbb{R}^{H \times W \times 1}, \quad
\left( \mathcal{Z}_s, \mathcal{Z}'_t, \mathcal{H} \right) \mapsto \Psi\left( \mathcal{Z}_s, \mathcal{Z}'_t, \mathcal{H} \right)=\mathcal{W}.
\label{eq:Psi}
\end{equation}

%%%%%%%%%%%%%%%%%%%%%%%%%%%%%%%%%%%%%%%%%%%%%%%%%%%%%%%%%%%%%%%%%%%%%%%%%%%%%%%%%%%
% Differentiable Optimizer
%%%%%%%%%%%%%%%%%%%%%%%%%%%%%%%%%%%%%%%%%%%%%%%%%%%%%%%%%%%%%%%%%%%%%%%%%%%%%%%%%%%
\subsection{Differentiable Optimizer}
\label{sec:differentiable_optimizer}

We introduce a differentiable optimizer $\Omega$ to estimate the deformation graph parameters $\mathcal{T}$, given the correspondence map $\mathcal{C}$, importance weights $\mathcal{W}$, and $N$ graph nodes $\mathcal{V}$:
\begin{equation}
\Omega: \mathbb{R}^{H \times W \times 2} \times \mathbb{R}^{H \times W \times 1} \times \mathbb{R}^{N \times 3} \rightarrow \mathbb{R}^{N \times 6}, \quad \left( \mathcal{C}, \mathcal{W}, \mathcal{V} \right) \mapsto \Omega\left( \mathcal{C}, \mathcal{W}, \mathcal{V} \right)=\mathcal{T},
\end{equation}
with $\mathcal{C}$ and $\mathcal{W}$ estimated by functions $\Phi$ (Eq.~\ref{eq:Phi}) and $\Psi$ (Eq.~\ref{eq:Psi}), respectively. 
Using the predicted dense correspondence map $\mathcal{C}$, we establish the data term for the non-rigid tracking optimization.
Specifically, we use a 2D data term that operates in image space and a depth data term that leverages the depth information of the input frames.
In addition to the data terms, we employ an As-Rigid-As-Possible regularizer~\cite{Sorkine2007arap} to encourage node deformations to be locally rigid, enabling robust deformation estimates even in the presence of noisy input cues. 
Note that the resulting optimizer module $\Omega$ is fully differentiable, but contains no learnable parameters.
In summary, we formulate non-rigid tracking as the following nonlinear optimization problem:
\begin{equation}
\underset{\mathcal{T}}{\argmin} \big(\lambda_\textrm{2D}  E_\textrm{2D}(\mathcal{T}) + \lambda_\textrm{depth} E_\textrm{depth}(\mathcal{T}) + \lambda_\textrm{reg} E_\textrm{reg}(\mathcal{T}) \big).
\label{eq:nonrigid_energy}
\end{equation}
%
%%% Optimization terms
\paragraph{2D reprojection term.}
Given the outputs of the dense correspondence predictor and weighting function, $\Phi\left( \mathcal{I}_s, \mathcal{I}_t \right)$ and $\Psi\left( \mathcal{Z}_s, \mathcal{Z}'_t, \mathcal{H} \right)$, respectively, we query for every pixel $\mathbf{u}$ in the source frame its correspondence $\mathbf{c}_\mathbf{u}$ and weight $w_{\mathbf{u}}$ to build the following energy term:
\begin{equation}
E_\textrm{2D}(\mathcal{T}) =  \sum_{\mathbf{u} \in \Pi_s} w_{\mathbf{u}}^2 \norm{\pi_\mathbf{c} (\mathrm{Q}(\mathbf{p}_\mathbf{u}, \mathcal{T})) - \mathbf{c}_\mathbf{u}}^2_2,
\end{equation}
where $\pi_\mathbf{c}: \mathbb{R}^3 \to \mathbb{R}^2, \quad \mathbf{p} \mapsto \pi_\mathbf{c}(\mathbf{p})$ is a perspective projection with intrinsic parameters $\mathbf{c}$ and $\mathbf{p}_\mathbf{u} = \pi^{-1}_{\mathbf{c}}(\mathbf{u}, d_\mathbf{u})$ as defined in Eq.~\ref{eq:backproject}.
Each pixel is back-projected to 3D, deformed using the current graph motion estimate as described in Eq. \ref{eq:Q} and projected onto the target image plane. 
The projected deformed location is compared to the predicted correspondence $\mathbf{c}_\mathbf{u}$.
\paragraph{Depth term.}
The depth term leverages the depth cues of the source and target images.
Specifically, it compares the $z$ components of a warped source point, i.e., $[\mathrm{Q}(\mathbf{p}_\mathbf{u}, \mathcal{T})]_z$, and a target point sampled at the corresponding location $\mathbf{c}_\mathbf{u}$ using bilinear interpolation:
\begin{equation}
E_\textrm{depth}(\mathcal{T}) =  \sum_{\mathbf{u} \in \Pi_s} w_\mathbf{u}^2 \big( [\mathrm{Q}(\mathbf{p}_\mathbf{u}, \mathcal{T})]_z - [\mathrm{P}_t ( \mathbf{c}_\mathbf{u})]_z \big)^2.
\end{equation}
\paragraph{Regularization term.}
We encourage the deformation of neighboring nodes in the deformation graph to be locally rigid.
Each node $\mathbf{v}_i \in \mathcal{V}$ has at most $K=8$ neighbors in the set of edges $\mathcal{E}$, computed as nearest nodes using geodesic distances. 
The regularization term follows \cite{Sorkine2007arap}:
\begin{equation}
E_\textrm{reg}(\mathcal{T}) = \sum_{(\mathbf{v}_i, \mathbf{v}_j) \in \mathcal{E}}  \norm{ \mathrm{e}^{\widehat{\hspace{0pt}\boldsymbol{\omega}\hspace{0pt}}_{\mathbf{v}_i}} (\mathbf{v}_j - \mathbf{v}_i) + \mathbf{v}_i + \mathbf{t}_{\mathbf{v}_i} - (\mathbf{v}_j + \mathbf{t}_{\mathbf{v}_j}) }^2_2.
\end{equation}

%%% Gauss-Newton algorithm
Equation~\ref{eq:nonrigid_energy} is minimized using the Gauss-Newton algorithm, as described in Algorithm~\ref{alg:GN}.
In the following, we denote the number of correspondences by $|\mathcal{C}|$ and the number of graph edges by $|\mathcal{E}|$. Moreover, we transform all energy terms into a residual vector $\mathbf{r} \in \mathbb{R}^{3 |\mathcal{C}| + 3 |\mathcal{E}|}$.
For every graph node, we compute partial derivatives with respect to translation and rotation parameters, constructing a Jacobian matrix $\mathbf{J}~\in~\mathbb{R}^{(3 |\mathcal{C}| + 3 |\mathcal{E}|) \times 6 N}$, where $N$ is the number of nodes in the set of vertices $\mathcal{V}$.
Analytic formulas for partial derivatives are described in the supplemental material.

Initially, the deformation parameters are initialized to $\mathcal{T}_0 = \mathbf{0}$, corresponding to zero translation and identity rotations.
In each iteration $n$, the residual vector $\mathbf{r}_n$ and the Jacobian matrix $\mathbf{J}_n$ are computed using the current estimate $\mathcal{T}_n$, and the following linear system is solved (using LU decomposition) to compute an increment $\Delta \mathcal{T}$: 
\begin{equation}
\mathbf{J}_n^T \mathbf{J}_n \Delta \mathcal{T} = -\mathbf{J}_n^T \mathbf{r}_n.
\end{equation}
At the end of every iteration, the motion estimate $\mathcal{T}$ is updated as $\mathcal{T}_{n+1} = \mathcal{T}_n +  \Delta \mathcal{T}$.
Most operations are matrix-matrix or matrix-vector multiplications, which are trivially differentiable.
Derivatives of the linear system solve operation are computed analytically, as described in \cite{barron2016fast} and detailed in the supplement. 
We use $max\_iter = 3$ Gauss-Newton iterations, which encourages the correspondence prediction and weight functions, $\Phi$ and $\Psi$, respectively, to make predictions such that convergence in~3 iterations is possible.
In our experiments we use $(\lambda_{\textrm{2D}}, \lambda_{\textrm{depth}}, \lambda_{\textrm{reg}}) = (0.001, 1, 1)$.

\begin{algorithm}
\caption{Gauss-Newton Optimization}\label{alg:euclid}
\begin{algorithmic}[1]
\State $\mathcal{C}\gets \Phi\left( \mathcal{I}_s, \mathcal{I}_t \right)$ \Comment{Estimate correspondences}
\State $\mathcal{W}\gets \Psi\left( \mathcal{Z}_s, \mathcal{Z}'_t, \mathcal{H} \right)$ \Comment{Estimate importance weights}
\Function{Solver}{$\mathcal{C}, \mathcal{W}, \mathcal{V}$} 
\State $\mathcal{T}\gets \mathbf{0}$
\For{$n \gets 0$ to $max\_iter$}
    \State $\mathbf{J}, \mathbf{r} \gets \mathrm{ComputeJacobianAndResidual(\mathcal{V}, \mathcal{T}, \mathcal{Z}_s, \mathcal{Z}_t', \mathcal{C}, \mathcal{W})}$
    \State $\Delta \mathcal{T} \gets \mathrm{LUDecomposition}(\mathbf{J}^T \mathbf{J} \Delta \mathcal{T} = - \mathbf{J}^T \mathbf{r})$ \Comment{Solve linear system}
    \State $\mathcal{T} \gets \mathcal{T} + \Delta \mathcal{T}$ \Comment{Apply increment}
\EndFor\label{euclidendwhile}
\State \textbf{return} $\mathcal{T}$
\EndFunction
\end{algorithmic}
\label{alg:GN}
\end{algorithm}

%%%%%%%%%%%%%%%%%%%%%%%%%%%%%%%%%%%%%%%%%%%%%%%%%%%%%%%%%%%%%%%%%%%%%%%%%%%%%%%%%%%
% End-to-end Optimization
%%%%%%%%%%%%%%%%%%%%%%%%%%%%%%%%%%%%%%%%%%%%%%%%%%%%%%%%%%%%%%%%%%%%%%%%%%%%%%%%%%%
\subsection{End-to-end Optimization}
Given a dataset of samples $\mathcal{X}_{s, t} = \{ {[\mathcal{I}_s | \mathcal{P}_s], [\mathcal{I}_t | \mathcal{P}_t], \mathcal{V}} \}$, our goal is to find the parameters $\phi$ and $\psi$ of $\Phi_\phi$ and $\Psi_\psi$, respectively, so as to estimate the motion $\mathcal{T}$ of a deformation graph $\mathcal{G}$ defined over the source RGB-D frame.
This can be formulated as a differentiable optimization problem (allowing for back-propagation) with the following objective:
\begin{equation}
\underset{\phi, \psi}{\argmin} \sum_{\mathcal{X}_{s, t}} 
\lambda_{\mathrm{corr}}
\mathcal{L}_{\mathrm{corr}} (\phi)
+ 
\lambda_{\mathrm{graph}}
\mathcal{L}_{\mathrm{graph}} (\phi, \psi)
+
\lambda_{\mathrm{warp}}
\mathcal{L}_{\mathrm{warp}} (\phi, \psi)
\label{eq:optim}
\end{equation}

%%% Correspondence loss
\paragraph{Correspondence loss.} 
We use a robust $q$-norm as in \cite{sun2018pwc} to enforce closeness of correspondence predictions to ground-truth:
\begin{equation}
    \mathcal{L}_{\mathrm{corr}} (\phi) = \tilde{M}^{\mathcal{C}}
    ( \, | \Phi_{\phi} \left( \mathcal{I}_s, \mathcal{I}_t \right) - \tilde{\mathcal{C}} | + \epsilon)^q.
\end{equation}
Operator $|\cdot|$ denotes the $\ell_1$ norm, $q<1$ (we set it to $q=0.4$) and $\epsilon$ is a small constant. Ground-truth correspondences are denoted by $\tilde{\mathcal{C}}$. 
Since valid ground truth for all pixels is not available, we employ a ground-truth mask $\tilde{M}^{\mathcal{C}}$ to avoid propagating gradients through invalid pixels.

%%% Graph loss
\paragraph{Graph loss.}
We impose an $l_2$-loss on node translations $\mathbf{t}$ (ground-truth rotations are not available): 
\begin{equation}
    \mathcal{L}_{\mathrm{graph}} (\phi, \psi) = \tilde{M}^{\mathcal{V}} 
    \Big\lVert \big[ 
    \underbrace{\Omega \big(\Phi_\phi \left( \mathcal{I}_s, \mathcal{I}_t \right), \Psi_\psi \left( \mathcal{Z}_s, \mathcal{Z}'_t, \mathcal{H} \right), \mathcal{V} \big)}_{\mathcal{T}}
    \big]_{\mathbf{t}} - \tilde{\mathbf{t}} \Big\rVert^2_2,
    \label{eq:l_graph}
\end{equation}
where $[\,\cdot\,]_\mathbf{t} : \mathbb{R}^{N \times 6} \to \mathbb{R}^{N \times 3}, \quad \mathcal{T} \mapsto [\mathcal{T}]_\mathbf{t} = \mathbf{t}$ extracts the translation part from the graph motion~$\mathcal{T}$.
Node translation ground-truth is denoted by $\tilde{\mathbf{t}}$ and $\tilde{M}^{\mathcal{V}}$ masks out invalid nodes. 
Please see the supplement for further details on how $\tilde{M}^{\mathcal{V}}$ is computed.

%%% Warp loss
\paragraph{Warp loss.}
We have found that it is beneficial to use the estimated graph deformation $\mathcal{T}$ to deform the dense source point cloud $\mathcal{P}_s$ and enforce the result to be close to the source point cloud when deformed with the ground-truth scene flow $\tilde{\mathcal{S}}$:
\begin{equation}
\mathcal{L}_\textrm{warp} ( \phi, \psi ) = \tilde{M}^{\mathcal{S}} \Big\lVert \mathrm{Q} \Big(\mathcal{P}_s, \underbrace{\Omega \big(\Phi_\phi \left( \mathcal{I}_s, \mathcal{I}_t \right), \Psi_\psi \left( \mathcal{Z}_s, \mathcal{Z}'_t, \mathcal{H} \right), \mathcal{V} \big)}_{\mathcal{T}} \Big) - (\mathcal{P}_s + \tilde{\mathcal{S}}) \Big\rVert^2_2.
\end{equation}
Here, we extend the warping operation $\mathrm{Q}$ (Eq.~\ref{eq:Q}) to operate on the dense point cloud $\mathcal{P}_s$ element-wise, and define $\tilde{M}^{\mathcal{S}}$ to mask out invalid points.

Note that We found it to be a more general notation to disentangle them (e.g., for scenarios where graph nodes are not sampled on the RGB-D frame).

%%%%%%%%%%%%%%%%%%%%%%%%%%%%%%%%%%%%%%%%%%%%%%%%%%%%%%%%%%%%%%%%%%%%%%%%%%%%%%%%%%%
% Non-rigid Reconstruction Framework
%%%%%%%%%%%%%%%%%%%%%%%%%%%%%%%%%%%%%%%%%%%%%%%%%%%%%%%%%%%%%%%%%%%%%%%%%%%%%%%%%%%
\subsection{Neural Non-rigid Tracking for 3D Reconstruction} 
\label{sec:reco}
We introduce our differentiable tracking module into the non-rigid reconstruction framework of \citet{newcombe2015dynamicfusion}.
In addition to the dense depth ICP correspondences employed in the original method, which help towards local deformation refinement, we employ a keyframe-based tracking objective.
Without loss of generality, every 50th frame of the sequence is chosen as a keyframe, to which we establish dense correspondences including the respective weights.
We apply a conservative filtering of the predicted correspondences based on the predicted correspondence weights using a fixed threshold $\delta=0.35$ and re-weight the correspondences based on bi-directional consistency, i.e., keyframe-to-frame and frame-to-keyframe.
Using the correspondence predictions and correspondence weights of valid keyframes ($>50\%$ valid correspondences), the non-rigid tracking optimization problem is solved.
The resulting deformation field is used to integrate the depth frame into the canonical volume of the object.
We refer to the original reconstruction paper~\cite{newcombe2015dynamicfusion} for details regarding the fusion process.
\section{Experiments}

In the following, we evaluate our method quantitatively and qualitatively on both non-rigid tracking and non-rigid reconstruction.
To this end, we use the DeepDeform dataset~\cite{bozic2020deepdeform} for training, with the given $340$-$30$-$30$ train-val-test split of RGB-D sequences.
Both non-rigid tracking and reconstruction are evaluated on the hidden test set of the DeepDeform benchmark.

\begin{figure*}
\centering{
    \def\svgwidth{1\textwidth}
    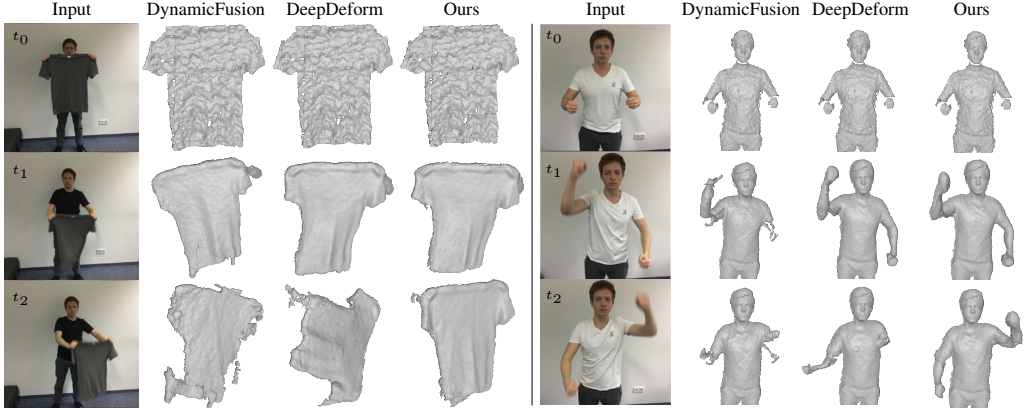
  \caption{Qualitative comparison of our method with DynamicFusion~\cite{newcombe2015dynamicfusion} and DeepDeform~\cite{bozic2020deepdeform} on test sequences from~\cite{bozic2020deepdeform}. The rows show different time steps of the sequence.}
  \label{fig:qualitative-comparison-svg}
}
\end{figure*}

\begin{table}[bp]
  \caption{
  We evaluate non-rigid tracking on the DeepDeform dataset \cite{bozic2020deepdeform}, showing the benefit of end-to-end differentiable optimizer losses and self-supervised correspondence weighting. 
  We denote correspondence prediction as $\Phi_{\textrm{c}}$, $\Phi_{\textrm{c+g}}$ and $\Phi_{\textrm{c+g+w}}$, depending on which losses $\mathcal{L}_\mathrm{corr}$, $\mathcal{L}_\mathrm{graph}$, $\mathcal{L}_\mathrm{warp}$ are used, and correspondence weighting as $\Psi_\textrm{supervised}$ and $\Psi_\textrm{self-supervised}$, either using an additional supervised loss or not.
  }
  \label{tab:tracking-results}
  \centering
  \begin{tabular}{lcc}
    \toprule
    Model & EPE 3D (\SI{}{\milli\meter}) & Graph Error 3D (\SI{}{\milli\meter}) \\
    \midrule
    $\Phi_{\textrm{c}}$ & $44.05$  & $67.25$  \\
    $\Phi_{\textrm{c+g}}$ & $39.12$  & $57.34$ \\
    $\Phi_{\textrm{c+g+w}}$ & $36.96$ & $54.24$ \\
    \midrule
    $\Phi_{\textrm{c}}$ + $\Psi_\textrm{supervised}$ & $28.95$  & $36.77$ \\
    $\Phi_{\textrm{c+g+w}}$ + $\Psi_\textrm{supervised}$ & $27.42$ & $34.68$   \\
    $\Phi_{\textrm{c+g+w}}$ + $\Psi_\textrm{self-supervised}$ & $\mathbf{26.29}$ & $\mathbf{31.00}$ \\
    \bottomrule
  \end{tabular}
\end{table}

%%%%%%%%%%%%%%%%%%%%%%%%%%%%%%%%%%%%%%%%%%%%%%%%%%%%%%%%%%%%%%%%%%%%%%%%%%%%%%%%%%%
% Implementation Details
%%%%%%%%%%%%%%%%%%%%%%%%%%%%%%%%%%%%%%%%%%%%%%%%%%%%%%%%%%%%%%%%%%%%%%%%%%%%%%%%%%%
\subsection{Training Scheme}
The non-rigid tracking module has been implemented using the PyTorch library \cite{pytorch2019} and trained using stochastic gradient descent with momentum $0.9$ and learning rate $10^{-5}$.
We use an Intel Xeon 6240 Processor and an Nvidia RTX 2080Ti GPU.
The parameters of the dense correspondence prediction network $\phi$ are initialized with a PWC-Net model pre-trained on FlyingChairs~\cite{dosovitskiy2015flownet} and FlyingThings3D~\cite{mayer16ft3d}.
We use a 10-factor learning rate decay every $10$k iterations, requiring about $30$k iterations in total for convergence, with a batch size of $4$.
For optimal performance, we first optimize the correspondence predictor $\Phi_\phi$ with $(\lambda_{\textrm{corr}}, \lambda_{\textrm{graph}}, \lambda_{\textrm{warp}}) = (5, 5, 5)$, without the weighting function $\Psi_\psi$.
Afterwards, we optimize the weighting function parameters $\psi$ with $(\lambda_{\textrm{corr}}, \lambda_{\textrm{graph}}, \lambda_{\textrm{warp}}) = (0, 1000, 1000)$, while keeping $\phi$ fixed.
Finally, we fine-tune both $\phi$ and $\psi$ together, with $(\lambda_{\textrm{corr}}, \lambda_{\textrm{graph}}, \lambda_{\textrm{warp}}) = (5, 5, 5)$.

\begin{table}[bp]
  \caption{
  Our method achieves state-of-the-art non-rigid reconstruction results on the DeepDeform benchmark \cite{bozic2020deepdeform}.
  Both our end-to-end differentiable optimizer and the self-supervised correspondence weighting are necessary for optimal performance.
  Not only does our approach achieve lower deformation and geometry error compared to state of the art, our correspondence prediction is about 85$\times$ faster.
  }
  \label{tab:reconstruction-results}
  \centering
  \begin{tabular}{lcc}
    \toprule
     Method    & Deformation error (\SI{}{\milli\meter}) &  Geometry error (\SI{}{\milli\meter}) \\
    \midrule
    DynamicFusion \cite{newcombe2015dynamicfusion} & $61.79$ & $10.78$ \\
    VolumeDeform \cite{innmann2016volumedeform} & $208.41$ & $74.85$ \\
    DeepDeform \cite{bozic2020deepdeform} & $31.52$ & $4.16$ \\
    \midrule
    Ours ($\Phi_{\textrm{c}}$)     & $54.85$ & $5.92$   \\
    Ours ($\Phi_{\textrm{c+g+w}}$)  & $53.27$ & $5.84$      \\
    Ours ($\Phi_{\textrm{c}}$ + $\Psi_\textrm{supervised}$)    & $40.21$ & $5.39$   \\
    Ours ($\Phi_{\textrm{c+g+w}}$ + $\Psi_\textrm{self-supervised}$)   &  $\mathbf{28.72}$ & $\mathbf{4.03}$   \\
    \bottomrule
  \end{tabular}
\end{table}

%%%%%%%%%%%%%%%%%%%%%%%%%%%%%%%%%%%%%%%%%%%%%%%%%%%%%%%%%%%%%%%%%%%%%%%%%%%%%%%%%%%
% Non-rigid Tracking Evaluation
%%%%%%%%%%%%%%%%%%%%%%%%%%%%%%%%%%%%%%%%%%%%%%%%%%%%%%%%%%%%%%%%%%%%%%%%%%%%%%%%%%%
\subsection{Non-rigid Tracking Evaluation}
For any frame pair $\mathcal{X}_{s,t}$ in the DeepDeform data \cite{bozic2020deepdeform}, we define a deformation graph $\mathcal{G}$ by uniformly sampling graph nodes $\mathcal{V}$ over the source object in the RGB-D frame, given a segmentation mask of the former.
Graph node connectivity $\mathcal{E}$ is computed using geodesic distances on a triangular mesh defined over the source depth map.
As a pre-processing step, we filter out any frame pairs where more than~$30\%$ of the source object is occluded in the target frame.
In Table~\ref{tab:tracking-results} non-rigid tracking performance is evaluated by the mean translation error over node translations $\mathbf{t}$ (Graph~Error~3D), where the latter are compared to ground-truth with an $l_2$ metric.
In addition, we evaluate the dense end-point-error (EPE 3D) between the source point cloud deformed with the estimated graph motion, $\mathrm{Q}(\mathcal{P}_s, \mathcal{T})$, and the source point cloud deformed with the ground-truth scene flow, $\mathcal{P}_s + \tilde{\mathcal{S}}$.
To support reproducibility, we report the mean error metrics of multiple experiments, running every setting~$3$~times.
We visualize the standard deviation with an error plot in the supplement.

We show that using graph and warp losses, $\mathcal{L}_\mathrm{graph}$ and $\mathcal{L}_\mathrm{warp}$, and differentiating through the non-rigid optimizer considerably improves both EPE 3D and Graph Error 3D compared to only using the correspondence loss $\mathcal{L}_\mathrm{corr}$.
Adding self-supervised correspondence weighting further decreases the errors by a large margin.
Supervised outlier rejection with binary cross-entropy loss does bring an improvement compared to models that do not optimize for the weighting function $\Psi_\psi$ (please see supplemental material for details on this supervised training of $\Psi_\psi$).
However, optimizing $\Psi_\psi$ in a \textit{self-supervised} manner clearly outperforms the former supervised setup.
This is due to the fact that, in the self-supervised scenario, gradients that flow from $\mathcal{L}_\mathrm{graph}$ and $\mathcal{L}_\mathrm{warp}$ through the differentiable solver $\Omega$ can better inform the optimization of $\Psi_\psi$ by minimizing the end-to-end alignment losses.

%%%%%%%%%%%%%%%%%%%%%%%%%%%%%%%%%%%%%%%%%%%%%%%%%%%%%%%%%%%%%%%%%%%%%%%%%%%%%%%%%%%
% Non-rigid Reconstruction Evaluation
%%%%%%%%%%%%%%%%%%%%%%%%%%%%%%%%%%%%%%%%%%%%%%%%%%%%%%%%%%%%%%%%%%%%%%%%%%%%%%%%%%%
\subsection{Non-rigid Reconstruction Evaluation}

We evaluate the performance of our non-rigid reconstruction approach on the DeepDeform benchmark \cite{bozic2020deepdeform} (see Table~\ref{tab:reconstruction-results}).
The evaluation metrics measure \emph{deformation error}, a 3D end-point-error between tracked and annotated correspondences, and \emph{geometry error}, which compares reconstructed shapes with annotated foreground object masks.
Our approach performs about $8.9\%$ better than the state-of-the-art non-rigid reconstruction approach of \citet{bozic2020deepdeform} on the deformation metric.
While our approach consistently shows better performance on both metrics, we also significantly lower the per-frame runtime to \SI{27}{\milli\second} per keyframe, in contrast to \cite{bozic2020deepdeform}, which requires \SI{2299}{\milli\second}.
Thus, our approach can also be used with multiple keyframes at interactive frames rates, e.g., \SI{90}{\milli\second} for $5$~keyframes and \SI{199}{\milli\second} for $10$~keyframes.

To show the influence of the different learned components of our method, we perform an ablation study by disabling either of our two main components: the end-to-end differentiable optimizer or the self-supervised correspondence weighting.
As can be seen, our end-to-end trained method with self-supervised correspondence weighting demonstrates the best performance.
Qualitatively, we show this in Figure~\ref{fig:qualitative-comparison-svg}.
In contrast to DynamicFusion~\cite{newcombe2015dynamicfusion} and DeepDeform~\cite{bozic2020deepdeform}, our method is notably more robust in fast motion scenarios.
Additional qualitative results and comparisons to the methods of \citet{guo2017real} and \citet{slavcheva2017killingfusion} are shown in the supplemental material.

\section{Conclusion}

We propose Neural Non-Rigid Tracking, a differentiable non-rigid tracking approach that allows learning the correspondence prediction and weighting of traditional tracking pipelines in an end-to-end manner.
The differentiable formulation of the entire tracking pipeline enables back-propagation to the learnable components, guided by a loss on the tracking performance.
This not only achieves notably improved tracking error in comparison to state-of-the-art tracking approaches, but also leads to better reconstructions, when integrated into a reconstruction framework like DynamicFusion~\cite{newcombe2015dynamicfusion}.
We hope that this work inspires further research in the direction of neural non-rigid tracking and believe that it is a stepping stone towards fully differentiable non-rigid reconstruction.
% Broader Impact
\section*{Broader Impact}

Our paper presents learned non-rigid tracking.
It is establishing the basis for the important research field of non-rigid tracking and reconstruction, which is needed for a variety of applications where man-machine and machine-environment interaction is required.
These applications range from the field of augmented and virtual reality to autonomous driving and robot control.
In the former, a precise understanding of dynamic and deformable objects is of major importance in~order to provide an immersive experience to the user.
Applications such as holographic calls would greatly benefit from research like ours.
This, in turn, could provide society with the next generation of 3D communication tools. 
On the other hand, as a low-level building block, our work has no direct negative outcome, other than what could arise from the aforementioned applications.
%

% Ackowledgments
\begin{ack}
This work was supported by the ZD.B (Zentrum Digitalisierung.Bayern), the Max Planck Center for Visual Computing and Communications (MPC-VCC), a TUM-IAS Rudolf M\"o{\ss}bauer Fellowship, the ERC Starting Grant Scan2CAD (804724), and the German Research Foundation (DFG) Grant Making Machine Learning on Static and Dynamic 3D Data Practical.
\end{ack}

{\small
\bibliographystyle{abbrvnat}
\bibliography{main}
}

%%%%%%%%%%%%%%%%%%%%%%%%%%%%
% Appendix
\newpage
\appendix

%%%%%%%%%%%%%%%%%%%%%%%%%%%%%%%%%%%%%%%%%%%%%%%%%%%%%%%%%%%%%%%%%%%%%%%%%%%%%%%%%%%
% Warping Operation
%%%%%%%%%%%%%%%%%%%%%%%%%%%%%%%%%%%%%%%%%%%%%%%%%%%%%%%%%%%%%%%%%%%%%%%%%%%%%%%%%%%
\section{Non-rigid Deformation Model}
\begin{figure*}[h]
\centering{
    \def\svgwidth{0.8\linewidth}
    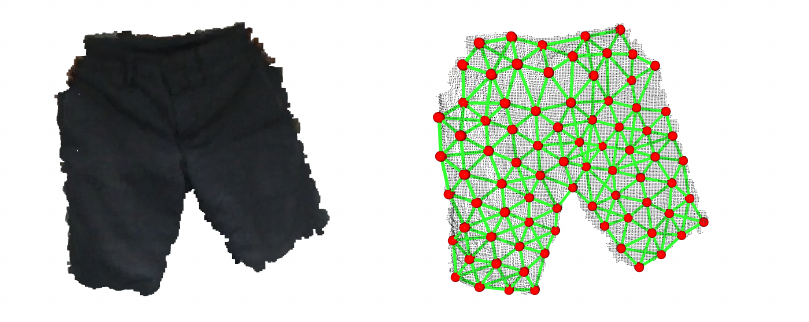
    \caption{Given a an object in the source RGB-D frame, we define a deformation graph $\mathcal{G}$ over the former. Nodes $\mathcal{V}$ (red spheres) are uniformly subsampled over the source RGB-D frame. Edges $\mathcal{E}$ (green lines) are computed between nodes based on geodesic connectivity among the latter.}
    \label{fig:object_and_graph}
}
\end{figure*}

To represent the dense motion from a source to a target RGB-D frame, we adapt the embedded deformation model of \citet{Sumner2007ed}.
We uniformly sample graph nodes $\mathcal{V}$ over the source RGB-D frame (see Fig.~\ref{fig:object_and_graph}), ensuring $\sigma$-coverage of the foreground object, i.e., the distance of every foreground point to the nearest graph node is at most $\sigma > 0$ (we set $\sigma = 0.05$~m).

For every node $v_i \in \mathcal{V}$, we estimate its global translation vector $\mathbf{t}_{\mathbf{v}_i} \in \mathbb{R}^3$ and local rotation matrix $\mathbf{R}_{\mathbf{v}_i} \in \mathbb{R}^{3 \times 3}$, represented in axis-angle notation as $\boldsymbol{\omega}_{\mathbf{v}_i} \in \mathbb{R}^3$.
Using the deformation parameters $\mathcal{T} = (\boldsymbol{\omega}_{\mathbf{v}_1}, \mathbf{t}_{\mathbf{v}_1}, \dots, \boldsymbol{\omega}_{\mathbf{v}_N}, \mathbf{t}_{\mathbf{v}_N})$ a 3D point $\mathbf{p} \in \mathbb{R}^{3}$ is deformed by interpolating the nodes' motion:
\begin{equation}
    \mathrm{Q} (\mathbf{p}, \mathcal{T}) = \sum_{\mathbf{v}_i \in \mathcal{V}}{ \alpha_{\mathbf{v}_i}^{\mathbf{p}} ( \mathrm{e}^{\widehat{\hspace{0pt}\boldsymbol{\omega}\hspace{0pt}}_{\mathbf{v}_i}} (\mathbf{p} - \mathbf{v}_i) + \mathbf{v}_i + \mathbf{t}_{\mathbf{v}_i}) }.
\end{equation}
The weights $\alpha_{\mathbf{v}_i}^{\mathbf{p}} \in \mathbb{R}$ are called \emph{skinning} weights and measure the influence of each node on the current point $\mathbf{p}$. They are computed as in DoubleFusion~\cite{DoubleFusion}:
\begin{equation*}
\alpha_{\mathbf{v}_i}^{\mathbf{p}} = C \mathrm{e}^{\cfrac{1}{2 \sigma^2} ||\mathbf{v}_i - \mathbf{p}||^2_2} .
\end{equation*}
Here, $C$ denotes the normalization constant, ensuring that skinning weights add up to one for point $\mathbf{p}$:
\begin{equation*}
\sum_{\mathbf{v}_i \in \mathcal{V}}{\alpha_{\mathbf{v}_i}^{\mathbf{p}}} = 1 .
\end{equation*}

For each node $\mathbf{v}_i \in \mathcal{V}$, we represent its rotation in axis-angle notation as $\boldsymbol{\omega}_{\mathbf{v}_i} \in \mathbb{R}^3$.
This representation has singularities for larger angles, i.e., two different vectors $\boldsymbol{\omega}$ and $\boldsymbol{\omega}'$ can represent the same rotation (for example keeping the same axis and increasing the angle by $2\pi$ results in identical rotation).
To avoid singularities, we decompose the rotation matrix into $\mathrm{e}^{\widehat{\hspace{0pt}\boldsymbol{\omega}\hspace{0pt}}_{\mathbf{v}_i}} = \mathrm{e}^{\widehat{\hspace{0pt}\boldsymbol{\epsilon}\hspace{0pt}}_{\mathbf{v}_i}} \mathbf{R}_{\mathbf{v}_i}$ with $\boldsymbol{\epsilon}_{\mathbf{v}_i} = 0$, therefore optimizing only for delta rotations that have rather small rotation angles.

%%%%%%%%%%%%%%%%%%%%%%%%%%%%%%%%%%%%
%%%%%%%%%%%%%%%%%%%%%%%%%%%%%%%%%%%%
%%%%%%%%%%%%%%%%%%%%%%%%%%%%%%%%%%%%

\section{Differentiable Non-rigid Optimization}

Our non-rigid optimization is based on the Gauss-Newton algorithm and minimizes an energy formulation that is based on three types of residual components: the 2D reprojection term, the depth term and the regularization term of the non-rigid deformation.

For a pixel $\mathbf{u} \in \Pi_s \subset \mathbb{R}^2$ and graph edge $(\mathbf{v}_i, \mathbf{v}_j) \in \mathcal{E}$, we define such terms as:
\begin{align*}
r_\textrm{2D}^{\mathbf{u}} (\mathcal{T}) &=  w_{\mathbf{u}} \big( \pi_\mathbf{c} (\mathrm{Q}(\mathbf{p}_\mathbf{u}, \mathcal{T})) - \mathbf{c}_\mathbf{u} \big) \\ 
r_\textrm{depth}^{\mathbf{u}} (\mathcal{T}) &= w_\mathbf{u} \big( [\mathrm{Q}(\mathbf{p}_\mathbf{u}, \mathcal{T})]_z - [\mathrm{P}_t ( \mathbf{c}_\mathbf{u})]_z \big) \\ 
r_\textrm{reg}^{\mathbf{v}_i, \mathbf{v}_j} (\mathcal{T}) &= \mathrm{e}^{\widehat{\hspace{0pt}\boldsymbol{\omega}\hspace{0pt}}_{\mathbf{v}_i}} (\mathbf{v}_j - \mathbf{v}_i) + \mathbf{v}_i + \mathbf{t}_{\mathbf{v}_i} - (\mathbf{v}_j + \mathbf{t}_{\mathbf{v}_j}),
\end{align*}
where $\mathbf{c}_\mathbf{u} \in \mathbb{R}^2$ and $w_{\mathbf{u}} \in \mathbb{R}$ represent the predicted correspondence and the importance weight, respectively; and $\mathbf{p}_\mathbf{u} = \pi^{-1}_{\mathbf{c}}(\mathbf{u}, d_\mathbf{u})$ is a 3D point corresponding to the pixel $\mathbf{u}$ with depth value $d_{\mathbf{u}}$.
The Gauss-Newton method is an iterative scheme.
In every iteration $n$, we compute the Jacobian matrix $\mathbf{J}_n$ and the residual vector $\mathbf{r}_n$, and get a solution increment $\Delta \mathcal{T}$ by solving the normal equations:
\begin{equation*}
\mathbf{J}_n^T \mathbf{J}_n \Delta \mathcal{T} = -\mathbf{J}_n^T \mathbf{r}_n.
\end{equation*}

The construction of the Jacobian matrix $\mathbf{J}\in \mathbb{R}^{(3 |\mathcal{C}| + 3 |\mathcal{E}|) \times 6N}$, consisting of partial derivatives of the residual vector $\mathbf{r}^{3 |\mathcal{C}| + 3 |\mathcal{E}|}$ with respect to deformation parameters $\mathcal{T}~=~(\boldsymbol{\omega}_{\mathbf{v}_1}, \mathbf{t}_{\mathbf{v}_1}, \dots, \boldsymbol{\omega}_{\mathbf{v}_N}, \mathbf{t}_{\mathbf{v}_N})~\in~\mathbb{R}^{6N}$ is detailed in Section~\ref{sec:jacobian}.
The linear system is solved using LU decomposition.
To enable differentiation through the entire Gauss-Newton solver, we have to ensure that the linear solve is differentiable. We detail the differentiable linear solve operation in Section~\ref{sec:lin_solver}.
%

%%%%%%%%%%%%%%%%%%%%%%%%%%%%%%%%%%%%%%%%%%%%%%%%%%%%%%%%%%%%%%%%%%%%%%%%%%%%%%%%%%%
% Jacobian Matrix for Gauss-Newton Optimization
%%%%%%%%%%%%%%%%%%%%%%%%%%%%%%%%%%%%%%%%%%%%%%%%%%%%%%%%%%%%%%%%%%%%%%%%%%%%%%%%%%%
\subsection{Partial Derivatives}
\label{sec:jacobian}
In the following, we derive the partial derivatives of the residual vector $\mathbf{r}$ with respect to $\boldsymbol{\epsilon}_{\mathbf{v}_i}$ and $\mathbf{t}_{\mathbf{v}_i}$ of every node $\mathbf{v}_i$, to construct the Jacobian matrix $\mathbf{J}$.
To simplify notation, we define the rotation operator that takes as input an angular velocity vector $\boldsymbol{\epsilon} \in \mathbb{R}^3$, rotation matrix $\mathbf{R} \in \mathbb{R}^{3 \times 3}$ and point $\mathbf{p} \in \mathbb{R}^3$ and outputs the rotated point:
\begin{equation*}
\mathrm{R} (\boldsymbol{\epsilon}, \mathbf{R}, \mathbf{p}) = \mathrm{e}^{\widehat{\hspace{0pt}\boldsymbol{\epsilon}\hspace{0pt}}} \mathbf{R} \mathbf{p}.
\end{equation*}
To compute the partial derivative with respect to $\boldsymbol{\epsilon}$, we follow the derivation from~\citet{blanco2010tutorial}:
\begin{equation*}
\left.\cfrac{\partial \mathrm{R} (\boldsymbol{\epsilon}, \mathbf{R}, \mathbf{p})}{\partial \boldsymbol{\epsilon}} \right|_{\boldsymbol{\epsilon} = 0} = - \widehat{\mathbf{R} \mathbf{p}}
\end{equation*}
Here, the $\, \widehat{\cdot}$-operator creates a $3 \times 3$ skew-symmetric matrix from a 3-dimensional vector. 

%
%%% Partial derivative of warping operation 
%
The rotation operator $\mathrm{R} (\boldsymbol{\epsilon}, \mathbf{R}, \mathbf{p})$ is a core part of the warping operator $\mathrm{Q} (\mathbf{p}, \mathcal{T})$. 
It follows that partial derivatives of a warping operator $\mathrm{Q} (\mathbf{p}, \mathcal{T})$ with respect to $\boldsymbol{\epsilon}_{\mathbf{v}_i}$ and $\mathbf{t}_{\mathbf{v}_i}$ for every node $\mathbf{v}_i$ can be computed as
\begin{align*}
\cfrac{\partial \mathrm{Q} (\mathbf{p}, \mathcal{T})}{\partial \boldsymbol{\epsilon}_{\mathbf{v}_i}} &= -\alpha_{\mathbf{v}_i}^{\mathbf{p}}  \reallywidehat{\mathbf{R}_{\mathbf{v_i}} (\mathbf{p} - \mathbf{v}_i)}, \\
\cfrac{\partial \mathrm{Q} (\mathbf{p}, \mathcal{T})}{\partial \mathbf{t}_{\mathbf{v}_i}} &= \alpha_{\mathbf{v}_i}^{\mathbf{p}} \mathbf{I}.
\end{align*}

%
%%% Partial derivative of intrinsic projection
%
Another building block of our optimization terms is the perspective projection $\pi_\mathbf{c}$ with intrinsic parameters $\mathbf{c} = (f_x, f_y, c_x, c_y)$:
\begin{align*}
\pi_\mathbf{c}&: \mathbb{R}^3 \rightarrow \mathbb{R}^2, \\
\pi_\mathbf{c} \left( 
\begin{bmatrix}
x \\
y \\
z
\end{bmatrix}
\right) &= 
\begin{bmatrix}
f_x \cfrac{x}{z} + c_x \\
f_y \cfrac{y}{z} + c_y
\end{bmatrix},
\end{align*}
whose partial derivatives with respect to the point $\mathbf{p} = (x, y, z)^\mathrm{T}$ are derived as
\begin{equation*}
\cfrac{\partial \pi_\mathbf{c} (\mathbf{p})}{\partial \mathbf{p}} = 
\begin{bmatrix}
\cfrac{f_x}{z} & 0 & -\cfrac{f_x x}{z^2} \\
0 & \cfrac{f_y}{z} & -\cfrac{f_y y}{z^2}
\end{bmatrix}.
\end{equation*}
By applying the chain rule, derivatives of all three optimization terms are computed.

%
%%% Partial derivatives of all optimization terms
%
\paragraph{Derivative of 2D reprojection term.}
For a pixel $\mathbf{u} \in \Pi_s \subset \mathbb{R}^2$ and its corresponding 3D point $\mathbf{p}_\mathbf{u}$, we derive partial derivatives of $r_\textrm{2D}^{\mathbf{u}} (\mathcal{T})$ as follows:
\begin{align*}
\cfrac{\partial r_\textrm{2D}^{\mathbf{u}} (\mathcal{T})}{\partial \boldsymbol{\epsilon}_{\mathbf{v}_i}} &= 
-w_{\mathbf{u}} \alpha_{\mathbf{v}_i}^{\mathbf{p}_\mathbf{u}}  
\begin{bmatrix}
\cfrac{f_x}{\mathbf{p}_\mathbf{u}^z} & 0 & -\cfrac{f_x \mathbf{p}_\mathbf{u}^x}{(\mathbf{p}_\mathbf{u}^z)^2} \\
0 & \cfrac{f_y}{\mathbf{p}_\mathbf{u}^z} & -\cfrac{f_y \mathbf{p}_\mathbf{u}^y}{(\mathbf{p}_\mathbf{u}^z)^2}
\end{bmatrix}
\reallywidehat{\mathbf{R}_{\mathbf{v_i}} (\mathbf{p}_\mathbf{u} - \mathbf{v}_i)}, \\ 
\cfrac{\partial r_\textrm{2D}^{\mathbf{u}} (\mathcal{T})}{\partial \mathbf{t}_{\mathbf{v}_i}} &= 
w_{\mathbf{u}} \alpha_{\mathbf{v}_i}^{\mathbf{p}_\mathbf{u}} 
\begin{bmatrix}
\cfrac{f_x}{\mathbf{p}_\mathbf{u}^z} & 0 & -\cfrac{f_x \mathbf{p}_\mathbf{u}^x}{(\mathbf{p}_\mathbf{u}^z)^2} \\
0 & \cfrac{f_y}{\mathbf{p}_\mathbf{u}^z} & -\cfrac{f_y \mathbf{p}_\mathbf{u}^y}{(\mathbf{p}_\mathbf{u}^z)^2}
\end{bmatrix}.
\end{align*}

\paragraph{Derivative of depth term.}
When computing the partial derivatives of the depth term $r_\textrm{depth}^{\mathbf{u}} (\mathcal{T})$, we need to additionally apply the projection to the $z$-component in the chain rule:
\begin{align*}
\cfrac{\partial r_\textrm{depth}^{\mathbf{u}} (\mathcal{T})}{\partial \boldsymbol{\epsilon}_{\mathbf{v}_i}} &= 
-w_{\mathbf{u}} \alpha_{\mathbf{v}_i}^{\mathbf{p}_\mathbf{u}} 
\begin{bmatrix}
0 & 0 & 1
\end{bmatrix}
\reallywidehat{\mathbf{R}_{\mathbf{v_i}} (\mathbf{p}_\mathbf{u} - \mathbf{v}_i)}, \\
\cfrac{\partial r_\textrm{depth}^{\mathbf{u}} (\mathcal{T})}{\partial \mathbf{t}_{\mathbf{v}_i}} &= 
w_{\mathbf{u}} \alpha_{\mathbf{v}_i}^{\mathbf{p}_\mathbf{u}} 
\begin{bmatrix}
0 & 0 & 1
\end{bmatrix}.
\end{align*}

\paragraph{Derivative of regularization term.}
For a graph edge $(\mathbf{v}_i, \mathbf{v}_j) \in \mathcal{E}$, the partial derivatives of $r_\textrm{reg}^{\mathbf{v}_i, \mathbf{v}_j} (\mathcal{T})$ with respect to $\boldsymbol{\epsilon}_{\mathbf{v}_i}$, $\mathbf{t}_{\mathbf{v}_i}$, $\boldsymbol{\epsilon}_{\mathbf{v}_j}$, $\mathbf{t}_{\mathbf{v}_j}$ are computed as:
\begin{align*}
\cfrac{\partial r_\textrm{reg}^{\mathbf{v}_i, \mathbf{v}_j} (\mathcal{T})}{\partial \boldsymbol{\epsilon}_{\mathbf{v}_i}} &= 
-\reallywidehat{\mathbf{R}_{\mathbf{v_i}} (\mathbf{v}_j - \mathbf{v}_i)}
~,&
\cfrac{\partial r_\textrm{reg}^{\mathbf{v}_i, \mathbf{v}_j} (\mathcal{T})}{\partial \boldsymbol{\epsilon}_{\mathbf{v}_j}} &= 
\mathbf{0}, \\
\cfrac{\partial r_\textrm{reg}^{\mathbf{v}_i, \mathbf{v}_j} (\mathcal{T})}{\partial \mathbf{t}_{\mathbf{v}_i}} &= 
\mathbf{I}
~,&
\cfrac{\partial r_\textrm{reg}^{\mathbf{v}_i, \mathbf{v}_j} (\mathcal{T})}{\partial \mathbf{t}_{\mathbf{v}_j}} &= 
-\mathbf{I}.
\end{align*}
%

%%%%%%%%%%%%%%%%%%%%%%%%%%%%%%%%%%%%%%%%%%%%%%%%%%%%%%%%%%%%%%%%%%%%%%%%%%%%%%%%%%%
% Analytic Derivative of Linear Solve Operation
%%%%%%%%%%%%%%%%%%%%%%%%%%%%%%%%%%%%%%%%%%%%%%%%%%%%%%%%%%%%%%%%%%%%%%%%%%%%%%%%%%%
\subsection{Differentiable Linear Solve Operation}
\label{sec:lin_solver}
To simplify the notation, in the following we use $\mathbf{A} = \mathbf{J}_n^T \mathbf{J}_n$, $\mathbf{b} = -\mathbf{J}_n^T \mathbf{r}_n$ and $\mathbf{x} = \Delta \mathcal{T}$, which results in the linear system of the form
\begin{equation}
\mathbf{A} \mathbf{x} = \mathbf{b}.
\end{equation}
For matrix $\mathbf{A} \in \mathbb{R}^{6N \times 6N}$ and vectors $\mathbf{b} \in \mathbb{R}^{6N}$ and $\mathbf{x} \in \mathbb{R}^{6N}$ we define the linear solve operation as
\begin{equation}
\Lambda: \mathbb{R}^{6N \times 6N} \times \mathbb{R}^{6N}  \rightarrow \mathbb{R}^{6N}, \quad \left( \mathbf{A}, \mathbf{b} \right) \mapsto \mathbf{A}^{-1} \mathbf{b} = \mathbf{x}.
\end{equation}
In order to compute the derivative of the linear solve operation, we follow the analytic derivative formulation of \citet{barron2016fast}. 
If we denote the partial derivative of the loss $\mathcal{L}$ with respect to linear system solution $\mathbf{x}$ as $\frac{\partial \mathcal{L}}{\partial \mathbf{x}}$, we can compute the partial derivatives with respect to matrix $\mathbf{A}$ and vector $\mathbf{b}$ as:
\begin{equation}
\cfrac{\partial \mathcal{L}}{\partial \mathbf{b}} = \mathbf{A}^{-1} \cfrac{\partial \mathcal{L}}{\partial \mathbf{x}},
\qquad \qquad
\cfrac{\partial \mathcal{L}}{\partial \mathbf{A}} = \left( \mathbf{A}^{-1} \cfrac{\partial \mathcal{L}}{\partial \mathbf{x}}  \right) \mathbf{x}^\mathrm{T} = -\cfrac{\partial \mathcal{L}}{\partial \mathbf{b}} \mathbf{x}^\mathrm{T}.
\end{equation}
Thus, the computation of $\frac{\partial \mathcal{L}}{\partial \mathbf{b}}$ requires solving a linear system with matrix $A$.
To solve this system, we re-use the LU decomposition from the forward pass.

%%%%%%%%%%%%%%%%%%%%%%%%%%%%%%%%%%%%%%%%%%%%%%%%%%%%%%%%%%%%%%%%%%%%%%%%%%%%%%%%%%%
% Ablations
%%%%%%%%%%%%%%%%%%%%%%%%%%%%%%%%%%%%%%%%%%%%%%%%%%%%%%%%%%%%%%%%%%%%%%%%%%%%%%%%%%%
\subsection{Ablations}
\label{sec:design_choices}

We experimented with different design choices for our solver.

%%%%%%%%%%%%%%%%%%%%%%%%%%%%%%%%%%%%%%%%%
\paragraph{ARAP edge re-weighting.}
In non-rigid tracking, it is possible to weight ARAP terms for every graph edge differently, depending on the distance between the nodes.
In our method, we sample nodes uniformly on the surface, thus, all edges have similar length ($7.13 \pm 1.38$ \SI{}{\centi\meter}). 
Hence, edge re-weighting changes EPE 3D only marginally: $0.8\%$ lower EPE 3D and $1.7\%$ lower Graph Error~3D.

%%%%%%%%%%%%%%%%%%%%%%%%%%%%%%%%%%%%%%%%%
\paragraph{Nearest-neighbor vs. bilinear depth sampling.} 
When querying depth after predicting 2D correspondences, we found bilinear sampling to perform better, with $5.8\%$ lower EPE 3D and $6.29\%$ lower Graph Error 3D compared to nearest-neighbor sampling.

%%%%%%%%%%%%%%%%%%%%%%%%%%%%%%%%%%%%%%%%%
\paragraph{Influence of graph density.} 
We sample graph nodes %uniformly on the surface
with \SI{5}{\centi\meter} node coverage, which fits well with our setup of \SI{11}{\gibi\byte} for training. %Working on denser graphs is not possible due to computational limits.
Using coarser graphs, with \SI{10}{\centi\meter} and \SI{15}{\centi\meter} node coverage resulted in poorer performance:  $5.49\%$ and $8.33\%$ higher EPE 3D, as well as $7.34\%$ and $28.46\%$ higher Graph Error 3D, respectively.
In turn, the memory footprint on the GPU during training (with batch size 4) decreases with node coverage: \SI{10513}{\mebi\byte}, \SI{6153}{\mebi\byte} and \SI{5931}{\mebi\byte} for \SI{5}{\centi\meter}, \SI{10}{\centi\meter} and \SI{15}{\centi\meter} node coverage, respectively.

%%%%%%%%%%%%%%%%%%%%%%%%%%%%%%%%%%%%%%%%%
\paragraph{Number of optimization steps.} \
We empirically found 3 solver iterations to be the best compromise between performance and computational cost. 
Unrolling 3 solver iterations instead of 1 / 2, results in $24.9\%$ / $0.16\%$ lower EPE 3D and $22.7\%$ / $0.23\%$ lower Graph Error 3D.
Using 4 iterations only improves slightly with respect to 3 ($0.04\%$ lower EPE 3D, $0.03\%$ lower Graph Error 3D).
More than 4 does not change performance notably.

%%%%%%%%%%%%%%%%%%%%%%%%%%%%%%%%%%%%%%%%%
\paragraph{Warp loss as a superset of graph loss.} \
Note that since we sample graph nodes on depth maps, the graph loss (Eq.~15 in the paper) is in practice a subset of the warp loss (Eq.~16 in the paper).
However, we found it to be a more general notation to disentangle them.
For instance, this notation is helpful for scenarios where graph nodes are not sampled on the RGB-D frame.

\section{Losses}
In this section, we provide implementation details related to the training losses used for end-to-end optimization.
Following the architecture of \citet{sun2018pwc}, our correspondence prediction function $\Phi$ computes a hierarchy of correspondence predictions, instead of only the correspondences at the highest resolution.
These predictions are used to compute the correspondence loss in a coarse-to-fine fashion (see Section~\ref{sec:coarse2fine_corr}).
To achieve numerically stable optimization (Gauss-Newton solver) during training, a ground-truth mask is needed for the graph loss.
In Section~\ref{sec:stable_opt}, we detail how to ensure stable optimization by filtering invalid graph nodes.
In the ablation studies, we also included a comparison to a weight function $\Psi$ that is trained in a supervised manner (see Table~1 in the main paper).
Section~\ref{sec:supervised_train} details the training of this baseline using a supervised binary cross-entropy loss.
%

%%%%%%%%%%%%%%%%%%%%%%%%%%%%%%%%%%%%%%%%%%%%%%%%%%%%%%%%%%%%%%%%%%%%%%%%%%%%%%%%%%%
% Coarse-to-fine Correspondence Loss
%%%%%%%%%%%%%%%%%%%%%%%%%%%%%%%%%%%%%%%%%%%%%%%%%%%%%%%%%%%%%%%%%%%%%%%%%%%%%%%%%%%
\subsection{Coarse-to-fine Correspondence Loss}
\label{sec:coarse2fine_corr}
The design of our correspondence prediction function $\Phi$ follows the PWC-Net~\cite{sun2018pwc} architecture that predicts the correspondences in a coarse-to-fine manner.
Initially the correspondences are predicted at a coarse resolution of $10 \times 7$ px, and then refined to a resolution of $20 \times 14$ px, etc.
In total, there are $L = 5$ levels in the correspondence hierarchy, and the finest level predictions are used in the differentiable non-rigid optimization.
The correspondence loss is applied on every level $l$, by bilinear downsampling of the groundtruth correspondences $\tilde{\mathcal{C}}$ to a coarser resolution, resulting in $\tilde{\mathcal{C}}^l$.
Similarly, the ground-truth mask matrix $\tilde{M}^{\mathcal{C}}$ is downsampled to a coarser version $\tilde{M}^{\mathcal{C}^l}$, to avoid propagating gradients through invalid pixels. 
For each training sample $(\mathcal{I}_s, \mathcal{I}_t)$ and every level $l$, we therefore compute ground-truth correspondences $\tilde{\mathcal{C}}^l$ and the ground-truth mask matrix $\tilde{M}^{\mathcal{C}^l}$.
At every level $l$ the correspondence loss has the following form:
\begin{equation}
    \mathcal{L}_{\mathrm{corr}}^l (\phi) = \tilde{M}^{\mathcal{C}^l}
    ( \, | \Phi_{\phi}^l \left( \mathcal{I}_s, \mathcal{I}_t \right) - \tilde{\mathcal{C}}^l | + \epsilon)^q.
\end{equation}
With $q < 1$ (in our case $q = 0.4$) and $\epsilon$ being a small constant.

%%%%%%%%%%%%%%%%%%%%%%%%%%%%%%%%%%%%%%%%%%%%%%%%%%%%%%%%%%%%%%%%%%%%%%%%%%%%%%%%%%%
% Invalid Graph Node Filtering for Numerically Stable Optimization
%%%%%%%%%%%%%%%%%%%%%%%%%%%%%%%%%%%%%%%%%%%%%%%%%%%%%%%%%%%%%%%%%%%%%%%%%%%%%%%%%%%
\subsection{Numerically Stable Optimization}
\label{sec:stable_opt}
The non-rigid tracking optimization objective includes data (correspondence and depth) terms  and a regularization (ARAP) term.
If we only use regularization term, the optimization problem becomes ill-posed, since any rigid transformation of all graph nodes has no effect on the regularization term.
In order to satisfy memory limits, we do not use all pixel correspondences $\mathcal{C}$ at training time, but instead randomly sample $10$k correspondences.  
The edge set $\mathcal{E}$ of the deformation graph $\mathcal{G} = (\mathcal{V}, \mathcal{E})$ is computed by connecting each graph node with $K = 8$ nearest nodes, using geodesic distances on the depth map mesh as a metric.
This can lead to multiple disconnected graph components, i.e., different node clusters.
To ensure the optimization problem is well-defined, we ensure that we have a minimum number of correspondences in each node cluster.
In our experiments we filter out all node clusters with less than $2000$ correspondences.
This filtering has to be reflected in the loss computation. Thus, we define the mask matrix $\tilde{M}^{\mathcal{V}}$ to have zeros for nodes from filtered clusters, which prevents gradient back-propagation through invalidated graph nodes.
%

%%%%%%%%%%%%%%%%%%%%%%%%%%%%%%%%%%%%%%%%%%%%%%%%%%%%%%%%%%%%%%%%%%%%%%%%%%%%%%%%%%%
% Supervised Weight Network Baseline
%%%%%%%%%%%%%%%%%%%%%%%%%%%%%%%%%%%%%%%%%%%%%%%%%%%%%%%%%%%%%%%%%%%%%%%%%%%%%%%%%%%
\subsection{Supervised Weight Network Baseline}
\label{sec:supervised_train}
As a baseline, we introduce a model where we supervise the optimization of the weighting function~$\Psi$.
The ground-truth correspondence weighting $\tilde{\mathcal{W}} \in \mathbb{R}^{H \times W \times 1}$ for this supervision is generated by comparing current correspondence predictions $\mathcal{C}$ against the ground-truth correspondences $\tilde{\mathcal{C}}$.
We compare 3D distances between correspondences, using the target depth map $\mathcal{D}_t$ to query corresponding depth values.
A pixel in the ground-truth weighting $\tilde{\mathcal{W}}$ is assigned a $1$ or $0$ depending on the correspondence error.
Optimal performance was achieved by assigning $1$ to correspondences that are at most \SI{0.1}{\meter} away from groundtruth, and $0$ to correspondences that are at least \SI{0.3}{\meter} away from groundtruth, without propagating any gradient through remaining correspondence weights.
Binary cross-entropy loss is used to optimize $\Psi$ in this supervised setting.

\section{Reproducibility}

%%%%%%%%%%%%%%%%%%%%%%%%%%%%%%%%%%%%%%%%%%%%%%%%%%%%%%%%%%%%%%%%%%%%%%%%%%%%%%%%%%%
% Time and Memory Complexity
%%%%%%%%%%%%%%%%%%%%%%%%%%%%%%%%%%%%%%%%%%%%%%%%%%%%%%%%%%%%%%%%%%%%%%%%%%%%%%%%%%%
\subsection{Time and Memory Complexity}
\paragraph{Correspondence Prediction $\Phi$.}
Function $\Phi$ scales as a standard convolutional neural network linearly with the number of pixels in the input image (both in time and memory complexity).
The number of layers and kernel sizes are independent of the input an, thus, constant.
Our correspondence prediction network $\Phi$ consists of $55$ layers with a total number of $9.374$M parameters.
The processing of an image of resolution $640\times480$ takes $21.6$ ms.

\paragraph{Correspondence Weighting $\Psi$.}
Function $\Psi$ is a convolutional neural network with a fixed number of layers and kernels, and, thus, has a complexity that is linear in the number of pixels of the input image.
In total the network consists of $316$K learnable parameters that are distributed among $7$ layers.
For a forward pass with an image of resolution $640\times480$ the network takes $5.5$ ms.

\paragraph{Differentiable Optimizer.} 
Time and memory complexity of our differentiable Gauss-Newton solver is dominated by two operations: matrix-matrix multiplication of $\mathbf{J}^\mathrm{T}$ and $\mathbf{J}$ and LU decomposition of $\mathbf{J}^\mathrm{T} \mathbf{J}$ matrix.
For a matrix $\mathbf{J} \in \mathbb{R}^{(3 |\mathcal{C}| + 3 |\mathcal{E}|) \times 6N}$ the matrix-matrix multiplication $\mathbf{J}^\mathrm{T} \mathbf{J}$ has a time complexity of $\mathrm{O} (N^2 \cdot (|\mathcal{C}| + |\mathcal{E}|))$ and memory complexity of $\mathrm{O} (N \cdot (|\mathcal{C}| + |\mathcal{E}|))$.
We denoted the number of correspondences with $|\mathcal{C}|$, the number of graph edges with $|\mathcal{E}|$ and the number of graph nodes with $N$.
On the other hand, the time and memory complexity of LU decomposition of a matrix $\mathbf{J}^\mathrm{T} \mathbf{J} \in \mathbb{R}^{6N \times 6N}$ is $\mathrm{O} (N^3)$ and $\mathrm{O} (N^2)$, respectively.
Note that LU is dominated by matrix-matrix multiplication; in theory there exist algorithms better then $n^3$, like $n^{2.376}$ based on the Coppersmith–Winograd algorithm~\cite{Coppersmith1990}.
The total time complexity is, therefore, $\mathrm{O} (N^2 \cdot (|\mathcal{C}| + |\mathcal{E}|) + N^3)$ and memory complexity is $\mathrm{O} (N \cdot (|\mathcal{C}| + |\mathcal{E}|) + N^2)$.
%

%%%%%%%%%%%%%%%%%%%%%%%%%%%%%%%%%%%%%%%%%%%%%%%%%%%%%%%%%%%%%%%%%%%%%%%%%%%%%%%%%%%
% Training Details
%%%%%%%%%%%%%%%%%%%%%%%%%%%%%%%%%%%%%%%%%%%%%%%%%%%%%%%%%%%%%%%%%%%%%%%%%%%%%%%%%%%
\vspace{-0.15cm}
\subsection{Training Details} 
\begin{figure}
\centering{
    \def\svgwidth{1.0\linewidth}
    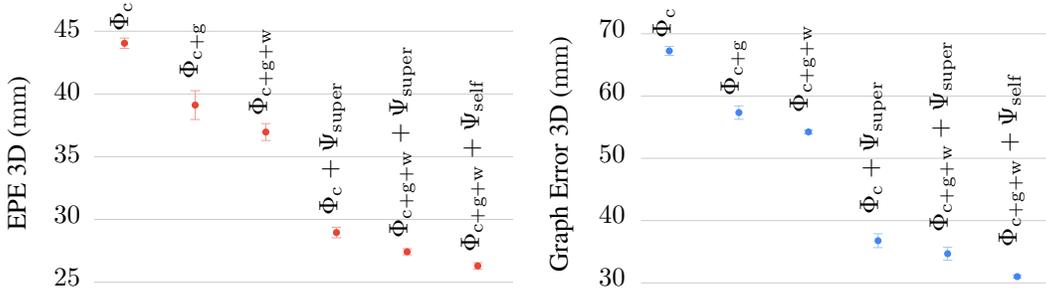
    \vspace{-0.3cm}
    \caption{Plots of non-rigid tracking EPE 3D (left) and Graph Error 3D (right) values, together with standard deviation bars. Corresponds to Table 1 in the main paper.}
    \label{fig:error_plots}
    \vspace{-0.3cm}
}
\end{figure}

\vspace{-0.05cm}
For reproducibility, the analysis of the achieved performance of the network with different training runs is important.
In the main paper (Table 1), we show an ablation study of our method and report average test errors of $3$ training runs.
In Figure~\ref{fig:error_plots} the corresponding standard deviations are plotted.
As can be seen, the training of our network is stable and results in small variations in performance.
For all experiments we used an Intel Xeon 6240 Processor with 18 cores and an Nvidia GeForce RTX 2080Ti GPU.
A typical power consumption of Nvidia 2080Ti GPU is around $280$ Watts.
Network experiments were run for $30$k iterations with batch size $4$, requiring in total about $14$ hours till convergence.

%%%%%%%%%%%%%%%%%%%%%%%%%%%%%%%%%%%%%%%%%%%%%%%%%%%%%%%%%%%%%%%%%%%%%%%%%%%%%%%%%%%
% Correspondence Filtering at Non-rigid Reconstruction
%%%%%%%%%%%%%%%%%%%%%%%%%%%%%%%%%%%%%%%%%%%%%%%%%%%%%%%%%%%%%%%%%%%%%%%%%%%%%%%%%%%
\vspace{-0.15cm}
\subsection{Keyframe-based Non-rigid Reconstruction} 
\vspace{-0.05cm}
To achieve robust non-rigid reconstruction, we propose the usage of a keyframe-based strategy.
We explored different keyframe sampling, as shown in Table~\ref{tab:less_keyframes}, and opted for sampling a keyframe every 50 frames in our setup.
For every keyframe, the correspondences to the latest frame in the video are predicted. 
Our neural non-rigid tracker provides us with correspondence $\mathbf{c}_\mathbf{u}$ and an importance weight $w_\mathbf{u} \in [0, 1]$ for every pixel $\mathbf{u} \in \Pi_s \subset \mathbb{R}^2$. 
We invalidate all correspondences of a keyframe with $w_\mathbf{u} < \delta$ (we set $\delta = 0.35$ in all experiments).
In case of large occlusions between the current frame and the keyframe, many correspondences are invalid.
If $50\%$ of the correspondences are invalid, we completely ignore the keyframe, which leads to less outliers and faster runtime.

\begin{table}[h]
  \caption{
  We evaluate how the deformation error (mm) varies with the keyframe sampling. More frames means lower keyframe sampling rate, i.e., larger frame-to-frame motion.
  }
  \label{tab:less_keyframes}
  \centering
  \begin{tabular}{lc}
    \toprule
    Keyframe density & Deformation error (\SI{}{\milli\meter}) \\
    \midrule
    DeepDeform \cite{bozic2020deepdeform} (filtering w/ neighboring frames)  & $31.52$ \\
    Ours: keyframe every 100 frames & $30.70$ \\
    Ours: keyframe every 75 frames & $29.68$ \\
    Ours: keyframe every 50 frames & $\mathbf{28.72}$ \\
    \bottomrule
  \end{tabular}
\end{table}

In addition, we apply correspondence reweighting based on cycle consistencies.
Specifically, we enforce bi-directional consistency and multi-keyframe consistency.
\textit{Bi-directional consistency} enables us to detect self-occlusions between a keyframe and current frame.
Correspondences are predicted in both directions keyframe-to-frame and frame-to-keyframe.
If following the correspondence in forward keyframe-to-frame and afterwards in backward frame-to-keyframe direction results in a 3D error larger than \SI{0.20}{\meter}, we reject the correspondence.
For \textit{multi-keyframe consistency}, multiple keyframe-to-frame predictions are estimated that correspond to the same 3D point in the canonical volume and the mean prediction value is computed.
If any of the predictions is more than \SI{0.15}{\meter} away from the mean value, we reject all correspondences for a given 3D canonical point.

\section{Benchmark Results}

\begin{figure*}
\centering{
    \def\svgwidth{1.0\linewidth}
    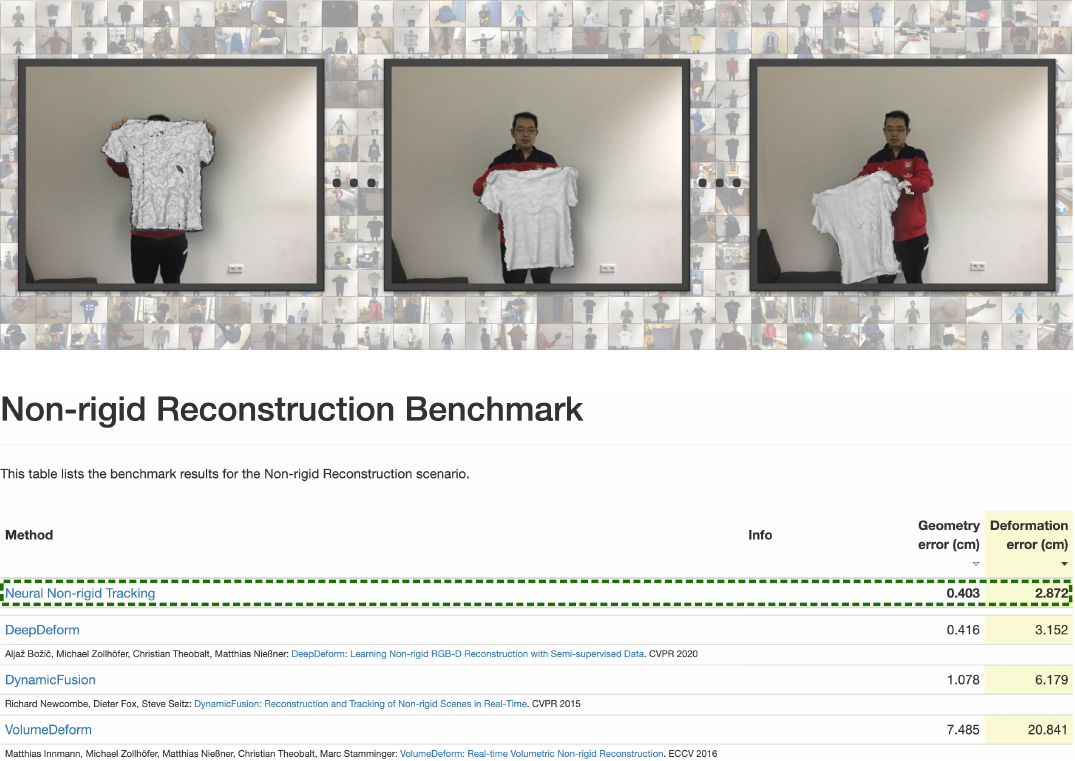
    \vspace{0.1cm}
    \caption{Screenshot of non-rigid reconstruction results on DeepDeform~\cite{bozic2020deepdeform} benchmark (taken on~11th~June~2020). }
    \label{fig:benchmark-screenshot}
}
\end{figure*}

In Figure~\ref{fig:benchmark-screenshot} we provide a screenshot of the currently best-performing, non-rigid reconstruction methods on the DeepDeform~\cite{bozic2020deepdeform} benchmark. Please visit \url{http://kaldir.vc.in.tum.de/deepdeform_benchmark/benchmark_reconstruction}.

\section{Additional Results}

\begin{figure*}
\centering{
    \def\svgwidth{1.0\linewidth}
    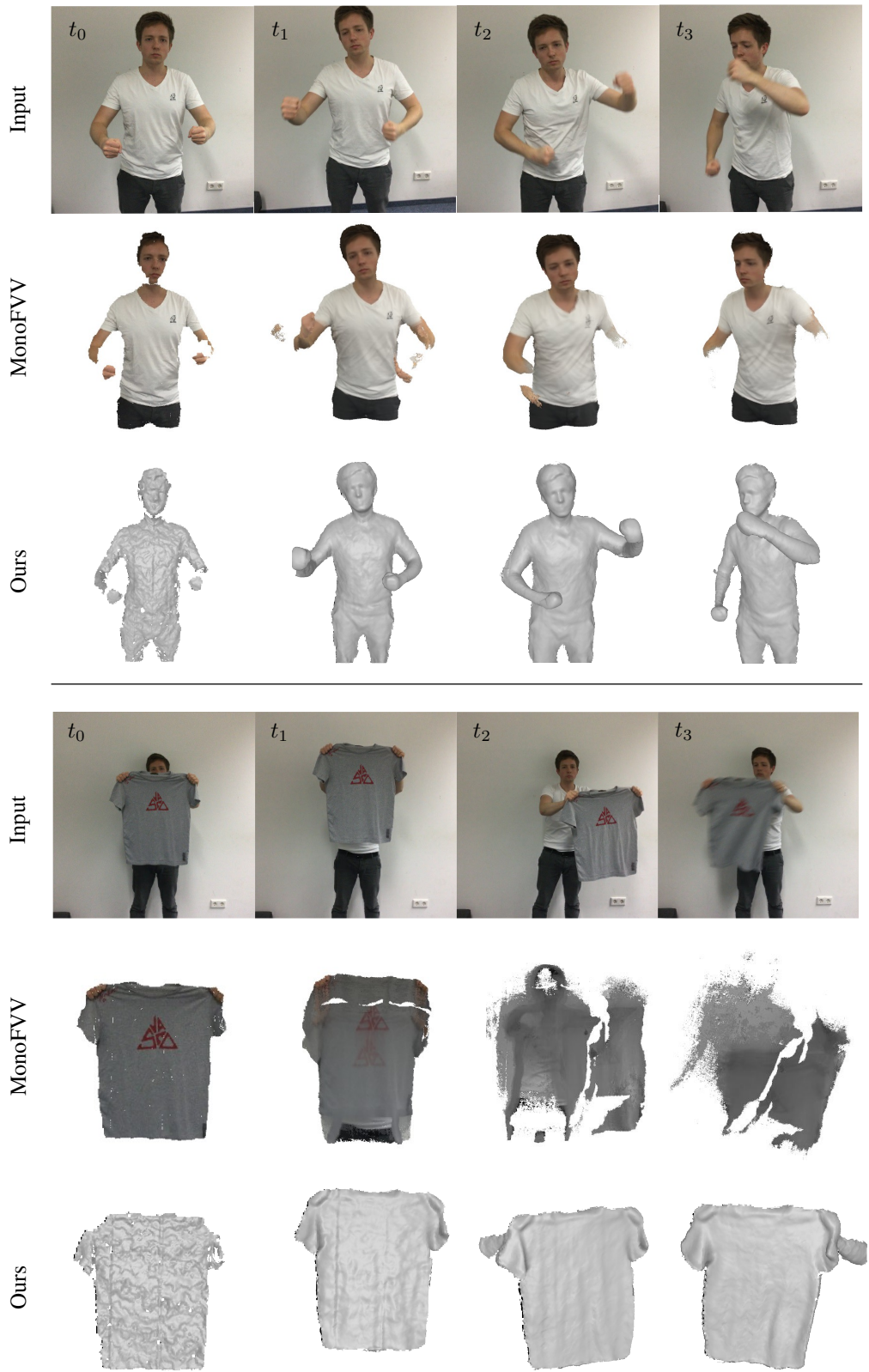
    \caption{Qualitative comparison of our method to MonoFVV~\cite{guo2017real} (test sequences from~\cite{bozic2020deepdeform}).}
    \label{fig:qualitative-comparison-monofvv}
}
\end{figure*}

In the following, we present additional qualitative results of our method in comparison to state-of-the-art methods.
Figure~\ref{fig:qualitative-comparison-monofvv} shows a comparison to \citet{guo2017real}.
As can be seen, our method better handles non-rigid movements with fast motions (t-shirt) and occlusions (arm).

\begin{figure*}
\centering{
    \def\svgwidth{0.94\linewidth}
    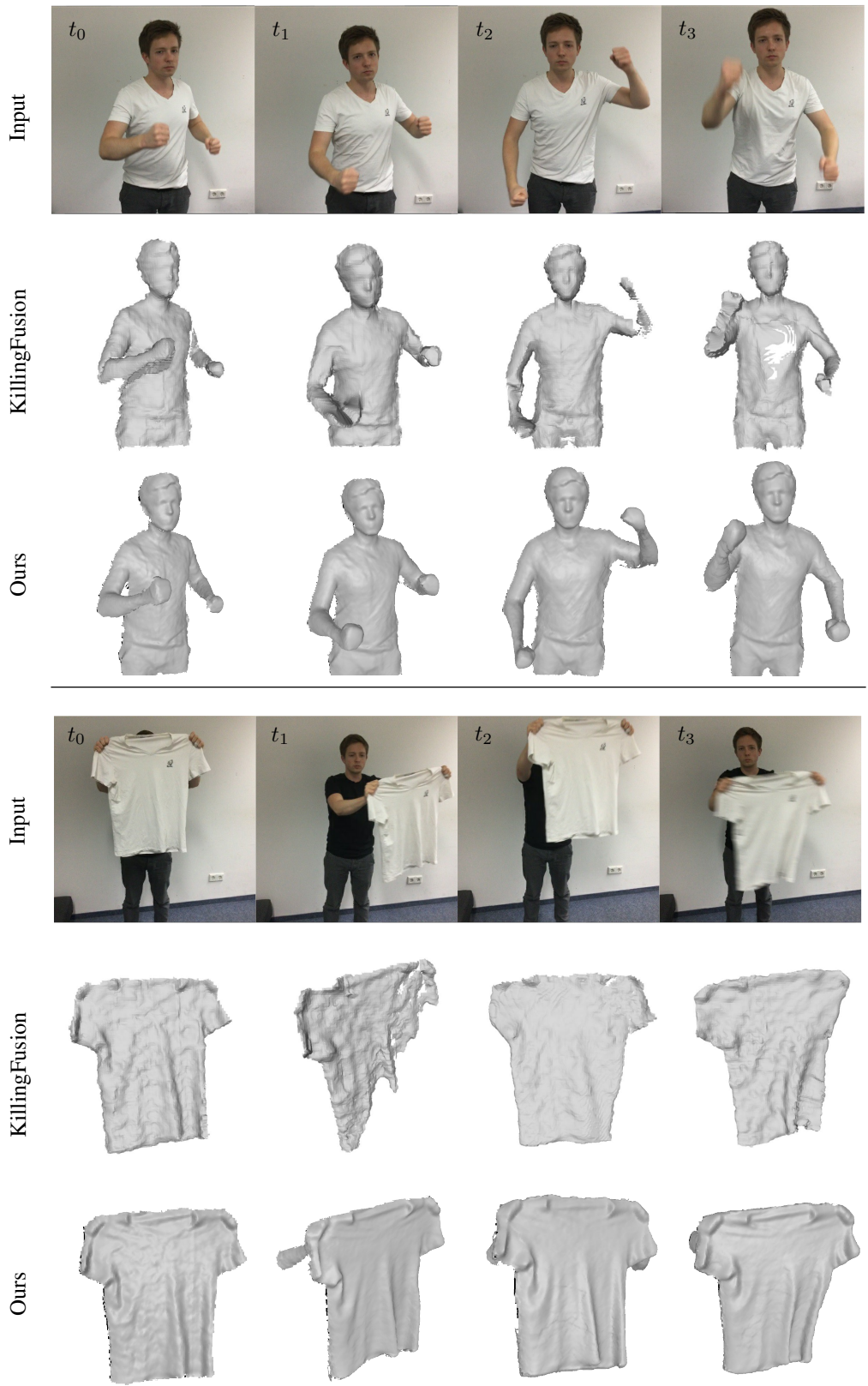
    \caption{Qualitative comparison of our method to KillingFusion~\cite{slavcheva2017killingfusion} (test sequences from~\cite{bozic2020deepdeform}).}
    \label{fig:qualitative-comparison-killingfusion}
}
\end{figure*}

In Figure~\ref{fig:qualitative-comparison-killingfusion}, we show the results of applying our method on test sequences of the DeepDeform dataset~\cite{bozic2020deepdeform}, and compare to the results of \citet{slavcheva2017killingfusion}. 
The reconstruction of our method leads to more complete and smooth meshes.
Note that the results of both methods \cite{guo2017real} and \cite{slavcheva2017killingfusion} were kindly provided by the authors.

\begin{figure*}
\centering{
    \def\svgwidth{0.95\linewidth}
    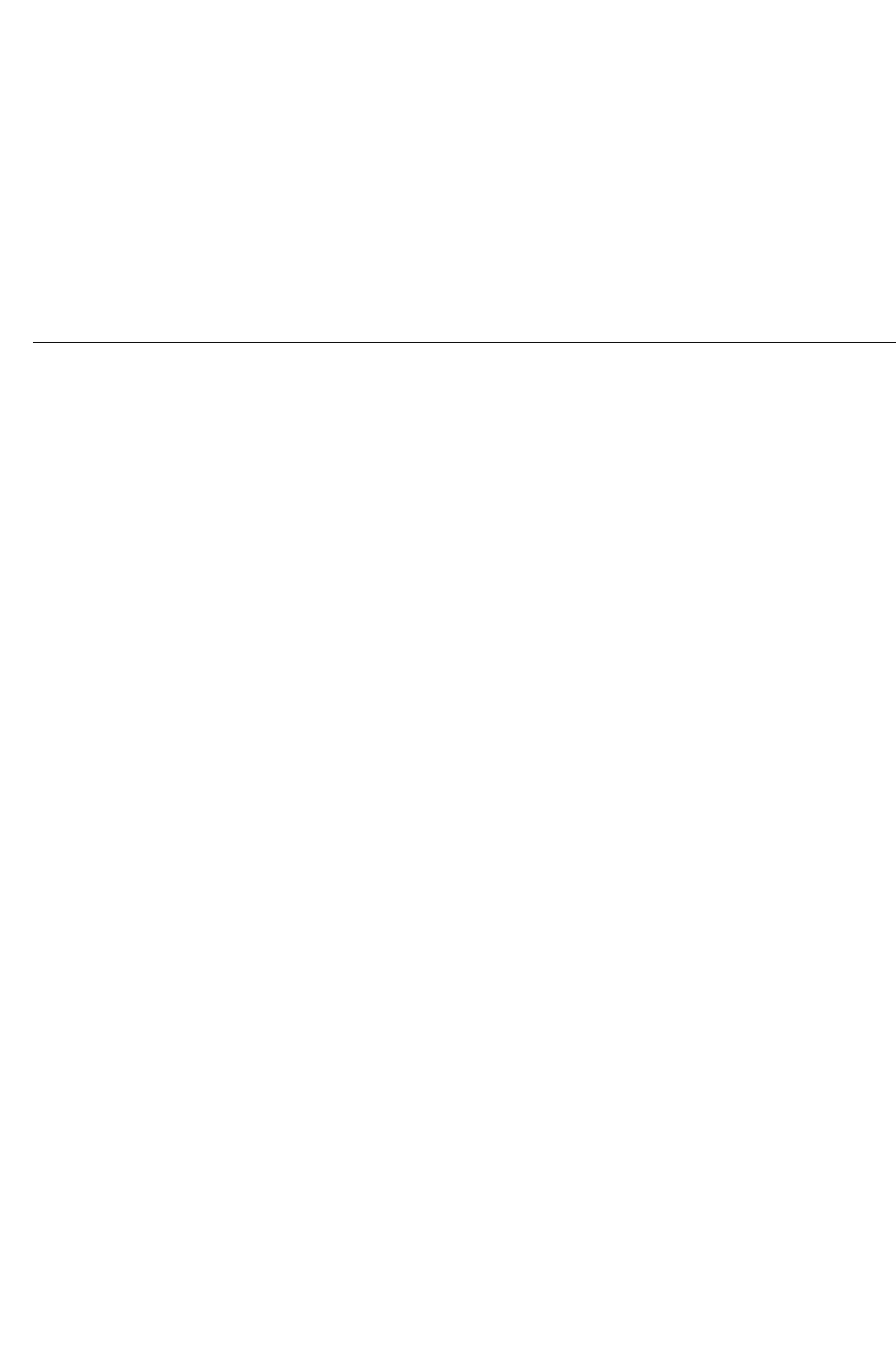
    \caption{Qualitative reconstruction results on VolumeDeform~\cite{innmann2016volumedeform} sequences.}
    \label{fig:qualitative-reconstruction-volumedeform-data}
}
\end{figure*}

We show qualitative reconstruction results of our method on VolumeDeform~\cite{innmann2016volumedeform} sequences in Figure~\ref{fig:qualitative-reconstruction-volumedeform-data}.
Our method can robustly reconstruct these RGB-D sequences, despite the fact that a Kinect sensor was used to record them, whereas our training data was obtained using a Structure IO sensor.
This shows that our network predictions can generalize to a different structured-light sensor input.

%%%%%%%%%%%%%%%%
% DOUBLE FUSION COMPARISON
We also compared to DoubleFusion \cite{DoubleFusion} and BodyFusion \cite{yu2017bodyfusion}, which focus on human body reconstruction by assuming human body prior.
We were able to compare on a sequence provided by \cite{yu2017bodyfusion}, and even without assuming any explicit shape or motion priors, we achieve competitive performance.
In particular, our method achieves an average tracking error of $\SI{0.0317}{\meter}$, while BodyFusion and DoubleFusion achieve $\SI{0.0227}{\meter}$ and $\SI{0.0221}{\meter}$, respectively.
Note that these methods cannot reconstruct non-human sequences, unlike our approach.

\begin{figure*}
\centering{
    \includegraphics[width=0.9\textwidth]{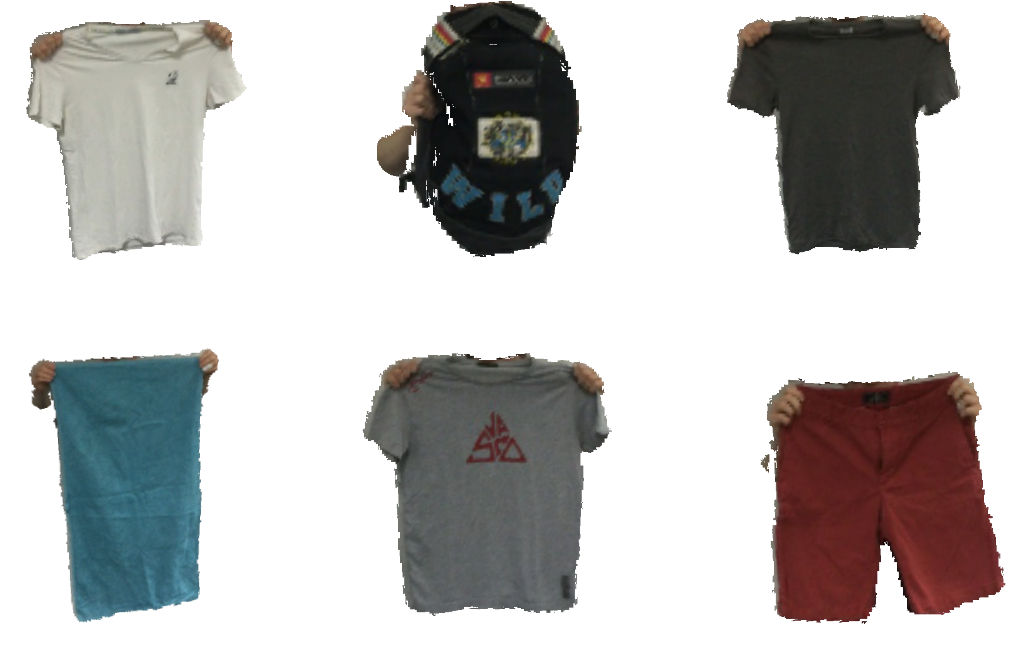}
    \caption{Texturing results, computed by aggregating color images over 100 frames of motion into a voxel grid.}
    \label{fig:textures}
}
\end{figure*}

%%%%%%%%%%%%%%%%
% Texture fusion
Additionally, in Figure \ref{fig:textures} we show texturing results, computed by aggregating color images over 100 frames of motion into a voxel grid.

%%%%%%%%%%%%%%%%%%%%%%%%%%%%

\end{document}